\documentclass[letterpaper,twocolumn,10pt]{article}
\usepackage{usenix-2020-09}

\usepackage{tikz}
\usepackage{amsmath}

\usepackage{filecontents}
\usepackage[]{hyperref}
\usepackage{blindtext}
\usepackage{caption}
\usepackage{subcaption}
\usepackage{amsmath}
\usepackage{amsfonts}
\usepackage{multirow}
\usepackage{tikz}
\usepackage{hyperref}
\usepackage{url}
\usepackage{algorithm}
\usepackage{algpseudocode}
\usepackage{booktabs}
\usepackage{threeparttable}
\usepackage{tabularx}
\usepackage{booktabs}
\usepackage{multirow}
\usepackage{graphicx}
\usepackage{wrapfig}

\newtheorem{thm}{\textbf{Observation}}
\usepackage[most]{tcolorbox}
\definecolor{light-gray}{gray}{0.95}
\tcolorboxenvironment{thm}{
   center,
   boxsep=0pt,
   top=0pt,
   bottom=0pt,
    width=\linewidth,
    colframe=light-gray,
    colback=light-gray
}

\newcommand{\sh}[1]{\textcolor{blue}{[??: #1]}}

\newcommand{\squishlist}{
\begin{list}{$\bullet$}
  { \setlength{\itemsep}{0pt}
     \setlength{\parsep}{0pt}
     \setlength{\topsep}{0pt}
     \setlength{\partopsep}{0pt}
     \setlength{\leftmargin}{0em}
     \setlength{\labelwidth}{0em}
     \setlength{\labelsep}{0.2em} } }
 
\newcommand{\squishlisttwo}{
\begin{list}{$\bullet$}
  { \setlength{\itemsep}{0pt}
     \setlength{\parsep}{0pt}
    \setlength{\topsep}{0pt}
    \setlength{\partopsep}{0pt}
    \setlength{\leftmargin}{2em}
    \setlength{\labelwidth}{1.5em}
    \setlength{\labelsep}{0.5em} } }
 
\newcommand{\squishend}{
  \end{list}  }

\newcommand{\DEL}[1]{\iffalse #1 \fi}

\begin{document}

\date{}

\title{\Large \bf HERA: Hybrid Edge-cloud Resource Allocation for Cost-Efficient AI Agents}

\author{
{\rm Shiyi Liu}\\
University of Virginia
\and
{\rm Haiying Shen}\\
University of Virginia
\and
{\rm Shuai Che}\\
Microsoft
\and
{\rm Mahdi Ghandi}\\
Microsoft
\and
{\rm Mingqin Li}\\
Microsoft
} 


\maketitle

\begin{abstract}
In the realm of AI, large language models (LLMs) like GPT-4, central to the operation of AI agents, predominantly operate in the cloud, incurring high operational costs. With local-based small language models (SLMs) becoming more accurate, the necessity of cloud-exclusive processing is being reconsidered. An AI agent's response to a user's request comprises a series of subtasks or iterations. Existing approaches only allocate a single request between SLM and LLM to ensure their outputs are similar, but adopting this approach in the AI agent scenario for assigning each subtask is not effective since SLM will output a different subsequent subtask, which affects the accuracy of the final output. In this paper, we first conduct experimental analysis to understand the features of AI agent operations. Leveraging our findings, we propose the \underline{H}ybrid \underline{E}dge-cloud \underline{R}esource \underline{A}llocation (HERA), a lightweight scheduler to automatically partition AI agent's subtasks between local-based SLM and cloud-based LLM. HERA considers the varying subtask features and strategically decides the location for each subtask in order to use SLM as much as possible while attaining the accuracy level. Our experimental results demonstrate that HERA increases accuracy by up to 9.1\% and SLM usage by up to 10.8\% compared to HybridLLM. It offloads 45.67\% of subtasks to a local SLM while attaining similar accuracy on average compared with the cloud-only LLM approach.
\end{abstract}

\section{Introduction}
The landscape of Natural language processing (NLP) has evolved with LLMs like GPT-4, showcasing unprecedented text generation capabilities \cite{ouyang2022training, wei2022emergent}. These models serve as the cognitive core in autonomous AI agents, revolutionizing traditional AI by integrating reasoning capability and tool use, thus enabling diverse interactions and generalization abilities \cite{sumers2023cognitive,handler2023balancing, wang2023survey, xi2023rise, nakano2021webgpt,yao2022react,lu2023chameleon, xu2023rewoo, wei2021finetuned}.  These agents, such as the notable Auto-GPT~\cite{AutoGPT} and AutoGen~\cite{wu2023autogen}, have expanded AI's reach. They try to achieve a given goal in natural language by breaking it into subtasks 
in an automatic process~\cite{yao2022react, mialon2023augmented, liang2023taskmatrix}. Each subtask represents a specific action or decision that the AI agent needs to perform to progress toward completing the overall user request. For example, when an AI agent receives a question ``Who was the maternal grandfather of the person who directed the 1997 film Titanic?'', it generates three subtasks ``Identify the director of Titanic (1997)'', ``Find the director's mother'', and ``Determine the mother's father''. 
The language model is invoked at each step to interpret the current state, execute the subtask, and generate the next subtask. This iterative process continues, with the agent potentially creating new subtasks or refining existing ones based on intermediate results, until it determines that the original request has been satisfactorily addressed. This process is reminiscent of advanced AI techniques like Chain of Thought (CoT) \cite{wei2022chain} and Reflective Processing \cite{shinn2023reflexion}, enhancing the agent's decision-making and problem-solving abilities. The effectiveness of these agents heavily depends on the language models' ability to accurately interpret and execute each subtask. 


Despite their capabilities, these LLM-based AI agents incur high operational costs (\$0.01/1K prompt tokens for GPT-4o) from frequent cloud-based API queries, presenting significant economic challenges \cite{Neoteric2023}. For example, incorporating Chat-GPT for enterprise use is expected to pose a financial burden exceeding \$9,000 monthly on small businesses \cite{Neoteric2023, exploding_topics_chatgpt_enterprise}. Previous work \cite {chen2023frugalgpt} tries to alleviate this problem by implementing open-sourced LLMs like LLAMA \cite{llama3}, Falcon \cite{almazrouei2023falcon} and OPT \cite{zhang2022opt}. However, the accuracy of open-sourced language models can be much worse on more complex tasks (i.e., requests) with multi-step reasoning in an AI agent scenario \cite{touvron2023llama, hsieh2023distilling}. If a company provides AI agent services using its own LLM APIs, it incurs high operational costs. Training and deploying LLMs can be prohibitively expensive, with costs ranging from millions of dollars for training\cite{brev_llm_cost_estimate}. 

To address these limitations, this work studies the computation breakdown for AI agents between the cloud and a local edge device motivated by previous findings that a SLM can achieve the accuracy of the LLM in some scenarios \cite{fu2023specializing, ranaldi2024aligning, xu2023small}. In particular, we investigate the possibility that some of the queries to LLM can be offloaded from the cloud to the local edge device. We discover that for certain subtasks of a user request, their simplicity allows for comparable accuracy when executed locally on a personal device instead of in the cloud. We also find that a static, independent allocation of each subtask to SLM or LLM throughout the AI agent process is not effective for avoiding accuracy degradation of the final output since SLM will output a different subsequent subtask compared to LLM. For the same reason, existing approaches like HybridLLM~\cite{ding2023hybrid}, which rely on a classifier to allocate a single task between SLM and LLM based on output similarity, are also ineffective when applied to subtask allocation in AI agent scenarios. For example, if an agent processes the question `Who was the maternal grandfather of the person who directed Titanic?', assigning the first subtask `Find the director of Titanic' to SLM might generate slightly different subsequent subtasks compared to using LLM, creating a cascading effect that impacts the final answer's accuracy. This interconnected nature of agent subtasks requires a more sophisticated approach that considers both the individual subtask characteristics and their position in the overall reasoning chain. A fine-grained subtask-level partitioning strategy based on the subtask and 
its position in the subtask sequence 
can potentially achieve higher SLM usage and comparable accuracy.

\begin{figure}
    \includegraphics[width=0.48\textwidth,height=\textheight,keepaspectratio]{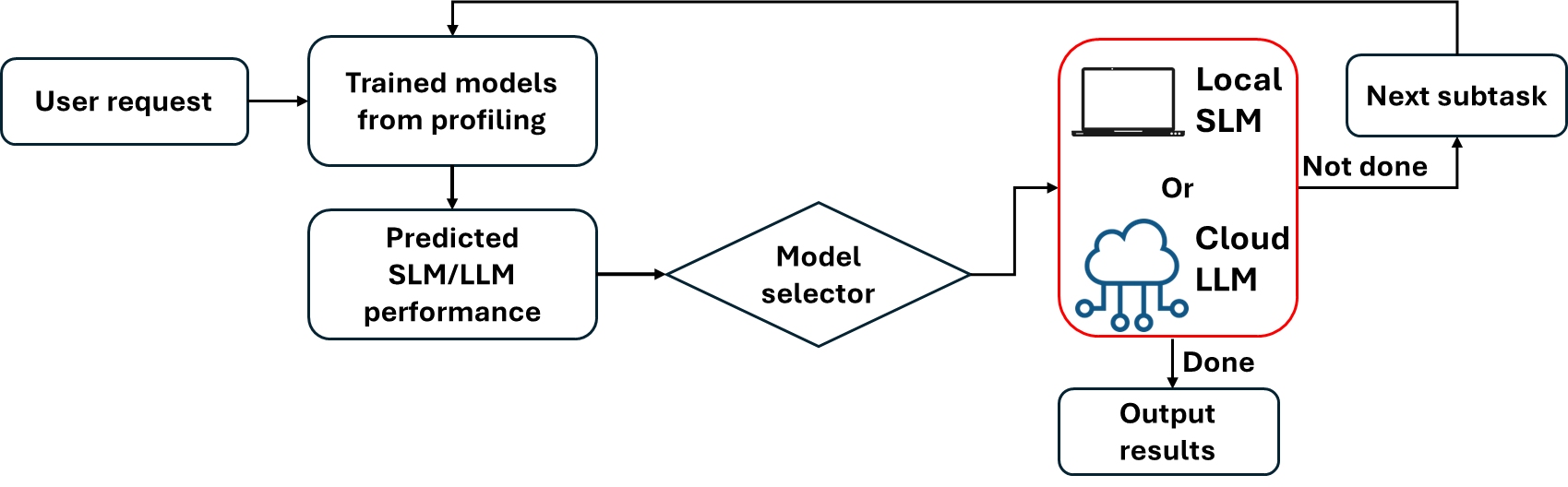}
    \caption{Dataflow of HERA. }
    \label{fig:overview}

\end{figure}

Based on these observations, we propose the \underline{A}daptive \underline{I}teration-level \underline{M}odel \underline{S}elector (HERA), a lightweight scheduler that moves AI agent computation at the subtask level from the cloud-based LLM to the local-based SLM as much as possible while preserving the accuracy. \textcolor{black}{HERA achieves a crucial balance by decreasing operational costs by up to 30\% through an allocation of 45.67\% of subtasks to local hardware, all while preserving accuracy within 2-5\% of cloud-exclusive approaches. Notably, despite using significantly less powerful local hardware compared to cloud infrastructure, HERA can achieve similar latency with using the cloud-only alternative. This makes HERA particularly valuable for organizations seeking to scale their AI agent deployments while managing costs. For a typical deployment processing 1 million requests monthly, HERA can reduce operational expenses by \$9,000-\$26,000 while maintaining high-quality service levels.}

As shown in Figure \ref{fig:overview}, 
HERA employs offline fine-tuned models to estimate subtask performance on SLM and LLM, and dynamically selects the most appropriate model for each subtask. It takes the following steps: 1) if the outputs from the SLM and LLM for the entire request are similar, SLM processes the whole request; 2) for a newly generated subtask, if the subtask's outputs from the SLM and LLM are similar, SLM processes the subtask; 3) if the outputs differ, HERA identifies a convergence point in the subtask series where the SLM and LLM outputs align and continues using the SLM until that point. 4) If no convergence is detected, the subtask is decomposed to facilitate processing by the SLM. Only when all the decomposed sub-subtasks can be handled by the SLM, they will be processed by the SLM. Otherwise, the original subtask will be processed by the LLM. 

\textcolor{black}{HERA differs from existing approaches like HybridLLM through its holistic treatment of AI agent subtasks. Rather than making isolated decisions, HERA recognizes the interconnected nature of agent reasoning and implements a hierarchical evaluation process. The system incorporates position-aware decision making, acknowledging that later subtasks require higher accuracy requirements. Furthermore, HERA introduces innovative features such as S-L distance metrics and convergence detection to optimize local processing opportunities. These architectural differences enable HERA to more effectively balance computational efficiency with accuracy across the entire chain of agent reasoning.}

Our contributions are as follows:
\begin{itemize}
\item \noindent\textbf{Experimental analysis on subtask allocation in AI agents.} We made several insightful observations from our experimental analysis. For example, unlike allocating a single request between the SLM and LLM that only needs to ensure the similarity of their outputs, allocating the subtasks for a request for AI agent has a unique challenge: if a subtask is allocated to SLM, the subsequent subtasks will vary, which affects
the accuracy of the final output. 


\item \noindent\textbf{Proposal of HERA.} Building on the observations from our experimental analysis, we propose HERA. HERA intelligently determines the allocation of subtasks between the local SLM and cloud-based LLM for AI agents in order
to use SLM as much as possible while attaining the accuracy level. 

\item \noindent\textbf{Comprehensive experiments of HERA.} HERA demonstrates superior performance, yielding a 9.1\% improvement in output accuracy and increases SLM usage by 10.8\% compared to HybridLLM. Additionally, HERA is able to offload 45.67\% of subtasks to a local SLM while maintaining comparable accuracy to the cloud-only LLM approach. 
\end{itemize}

This adaptive model selection framework, which is the first work in the realm of autonomous AI agents to our knowledge, represents a possibility of more efficient, cost-effective AI agent applications. 

\section{Motivation and Experiment Analysis}\label{sec2}

\textcolor{black}{In this section, we investigate the feasibility of executing AI agents partly on a local-based SLM and partly on a cloud-based LLM and compare with the status quo.}

 \subsection{Experiment Settings and Metrics} 
\label{sec:2.1}

We conducted experiments using the AutoGen framework on an Nvidia RTX 4090 GPU. We utilized two model pairs: Mistral-7B \cite{jiang2023mistral} with GPT-4, and Llama-3.1 8B \cite{touvron2023llama} with Claude 3.5 \cite{anthropic2023claude}. Unless otherwise indicated in the figure or table, we report the average results from these two pairs. Local models were run via llama.cpp \cite{llama.cpp}, while cloud models were accessed through APIs. The agentic framework is achieved by using AutoGen \cite{wu2023autogen} package autogen-agentchat 0.2. 

We evaluated on five datasets: GSM8K \cite{cobbe2021training}, HotPotQA \cite{yang2018hotpotqa}, DROP \cite{dua2019drop}, HumanEval \cite{chen2021evaluating}, and Webshop \cite{yao2022webshop}. These datasets are widely used for evaluating language model-based AI agents \cite{li2023making, wang2024adapting, wei2022chain, puerto2021metaqa, hu2024automated, shinn2023reflexion, yao2022react}. GSM8K has 8,500 grade school math word problems, and its accuracy is measured by the percentage of correct answers.
HotPotQA contains complex questions requiring reasoning.
DROP contains questions involving numerical operations and discrete reasoning. The accuracy of HotPotQA and DROP is measured by the F1 score.

HumanEval contains 164 programming problems, and its accuracy is measured by the pass@1 metric~\cite{chen2021codex}.
Webshop contains user purchasing requests and its accuracy is based on BERTScore \cite{zhang2019bertscore} compared to the ground-truth product descriptions. We consider a response to be similar if the BERTScore is higher than a threshold (e.g., 0.7). 
\emph{Completion rate} is the percentage of requests completed within 5 minutes.
\emph{Average number of subtasks} is the number of subtasks to complete a request. \emph{SLM usage} is the percentage of subtasks processed by SLM per request.
In this paper, we focus on text-based AI agents, therefore we use semantic similarity to quantify the alignment of SLM outputs and LLM outputs. We define two outputs as similar if the cosine similarity between their SBERT embeddings \cite{reimers2019sentence} exceeds a threshold. Unless otherwise specified, the threshold is set to 0.7, which is empirically determined. We report the average results from the two model pairs and for all datasets if not specifically indicated in the figure or table.

\subsection{SLM Usage for AI agent: Pros \& Cons}
We conducted experiments comparing the performance of five existing methods: HybridLLM~\cite{ding2023hybrid}, Oracle, All-SLM, All-LLM, and random assignment (Random). 
All-SLM processes an entire request using only the SLM, while All-LLM processes the request entirely using the LLM.
Oracle achieves the accuracy threshold (empirically determined as 90\% of the All-LLM's accuracy) while maximizing SLM usage by finding the optimal subtask assignment between the SLM and LLM for each user request. It is determined by enumerating all possible assignments for each subtask. 



\begin{figure}
    \centering
    \includegraphics[width=0.98\linewidth, keepaspectratio]{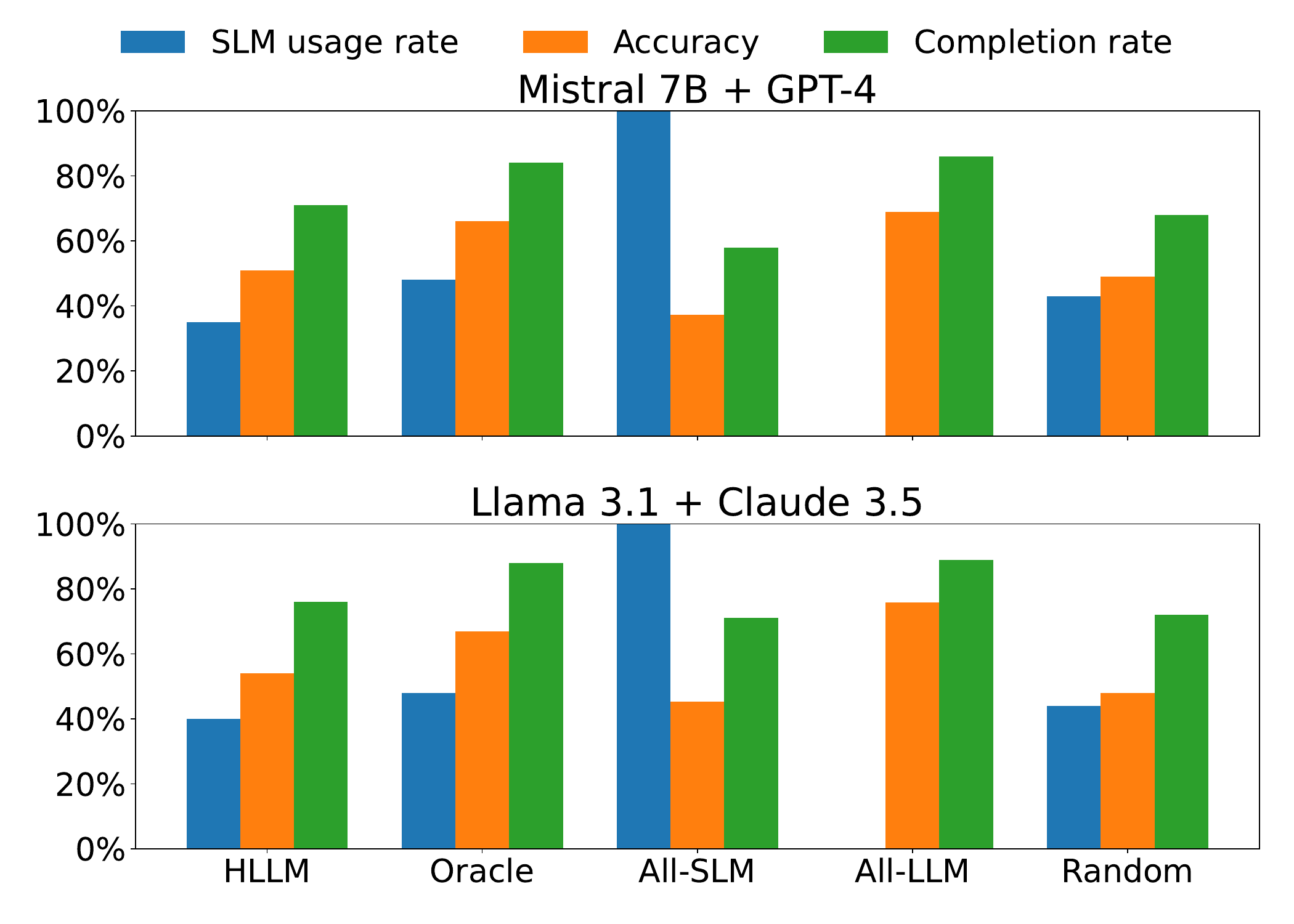}
    \caption{Performance of existing methods.}
    \label{fig:basic_compare}   
\end{figure}


Figure~\ref{fig:basic_compare} compares performance metrics across datasets and methods for two model pairs. Oracle consistently balances high accuracy and SLM usage, outperforming HybridLLM in both metrics. While All-SLM has the lowest accuracy and All-LLM the highest, Oracle slightly sacrifices accuracy for significantly increased SLM usage compared to All-LLM. Random allocation achieves moderate accuracy (51.3\%) and SLM usage (42.3\%). HybridLLM and Random's large accuracy drops may be impractical, whereas All-LLM is costly. Completion rates follow the order: Oracle $\approx$ All-LLM $>$ HybridLLM $>$ Random $>$ All-SLM. The consistent performance trends across model pairs suggest these results represent general patterns in hybrid LLM-SLM systems.


In addition, Table \ref{tab:incorrect_assignments} shows the percentage of incorrect assignment decisions of HybridLLM and Random compared to Oracle. An incorrect assignment occurs when a subtask is assigned to the SLM but Oracle assigns it to the LLM, or vice versa. HybridLLM shows over 35\% of incorrect assignments, while Random shows over 39\%. 
This is caused by independent assignment of individual subtasks without considering the interconnection between subtasks, implying the importance of holistic subtask model assignment.
The results highlight the suboptimality of the assignment decisions and suggests room for improvement in subtask allocation strategies. 

\begin{table}[h]
\caption{Percentage of incorrect assignments.}
\centering
\begin{tabular}{lcc}
\hline
\textbf{Method} & \textbf{Mistral+GPT-4} & \textbf{Llama+Claude 3.5} \\
\hline
HybridLLM & 35.2\% & 38.2\% \\
Random & 46.3\% & 39.2\% \\
\hline
\end{tabular}

\label{tab:incorrect_assignments}
\end{table}

\begin{thm}
\label{ob:1}
The existing HybridLLM inference system that assigns each subtask independently to either the SLM or LLM fails to maximize the accuracy or SLM usage in AI agent scenarios. Its performance gap from Oracle highlights the need for a more advanced approach. (Figure \ref{fig:basic_compare} and Table \ref{tab:incorrect_assignments})
\end{thm}



\subsection{Effects of Subtask-level Model Assignment}\label{sec2.4}



Building upon the above insights, we delved deeper into the performance differences between the SLM and LLM at the subtask level. In this experiment, 
we evaluated the accuracy impact of switching a single subtask from LLM to SLM while keeping the remaining subtasks in LLM. Conversely, we also assessed the scenario where all subtasks are executed in SLM, except for one that is switched to LLM. To account for user requests with varying numbers of subtasks, we grouped the subtasks into three relative positions: Early (first 1/3), Middle (middle 1/3), and Late (last 1/3) stages of the subtask sequence of a request.

\begin{wrapfigure}{r}{0.25\textwidth}
\includegraphics[width=0.98\linewidth, keepaspectratio]{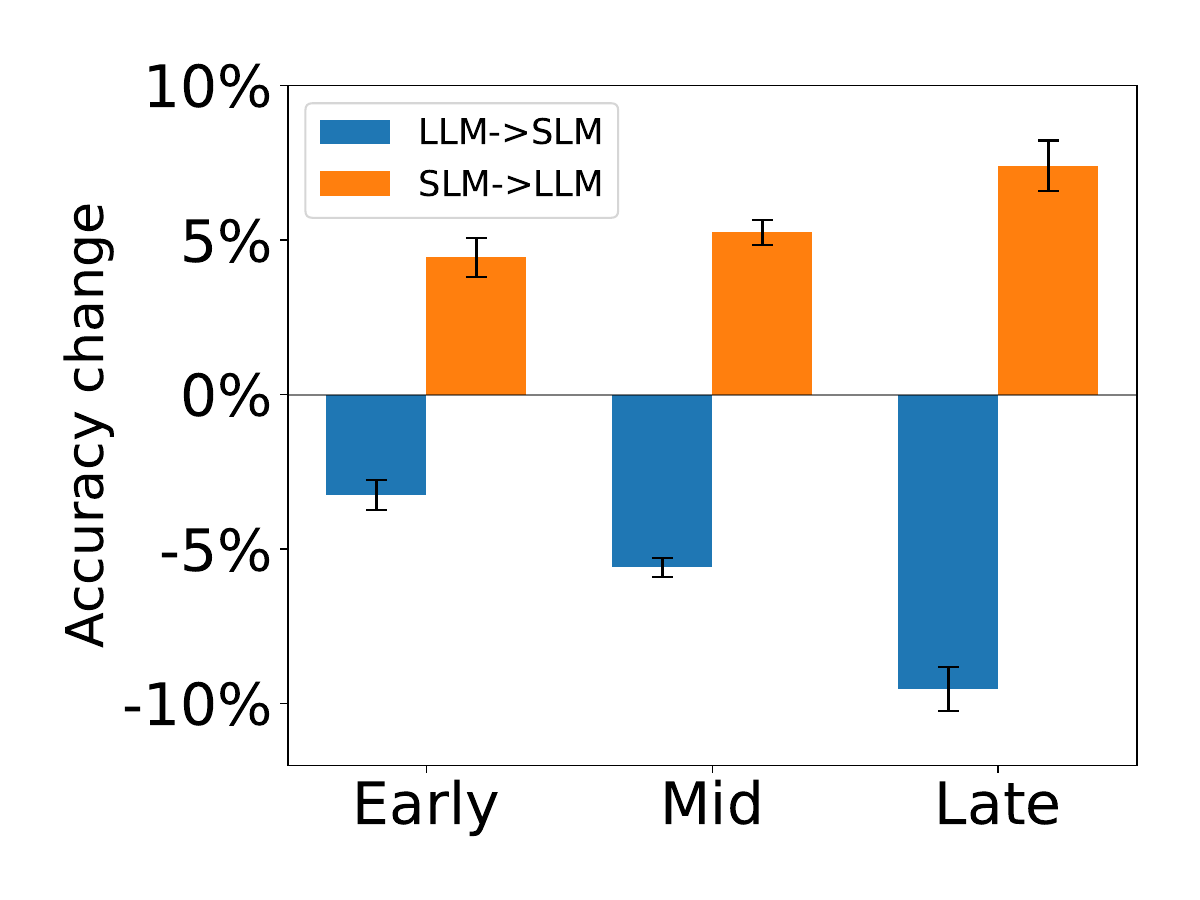} 
\caption{Accuracy changes by switching subtask between SLM and LLM.}
\label{fig:acc_change}
\end{wrapfigure}

Figure \ref{fig:acc_change} shows the average accuracy impact of switching a subtask at different stages. We switched each subtask from LLM to SLM and calculated the average change in accuracy at each stage. The same process was repeated for switching subtasks from SLM to LLM.

The results show that switching a subtask from LLM to SLM causes an average accuracy drop of 3.25\% in the Early stage, 5.59\% in the Middle stage, and 9.53\% in the Late stage. Conversely, switching from SLM to LLM yields accuracy gains of 4.44\%, 5.25\%, and 7.40\% in the Early, Middle, and Late stages, respectively. These findings suggest that SLM can manage early subtasks with minimal accuracy loss, but as tasks progress, leveraging LLM's advanced capabilities becomes increasingly critical.

\begin{thm} 
\label{ob:2} 

Subtask position can influences the accuracy impact of SLM-LLM switching, with later stages showing greater effects, highlighting its importance in allocation decisions.(Figure \ref{fig:acc_change})
\end{thm}

\begin{figure}
    \centering
    \includegraphics[width=0.98\linewidth, keepaspectratio]{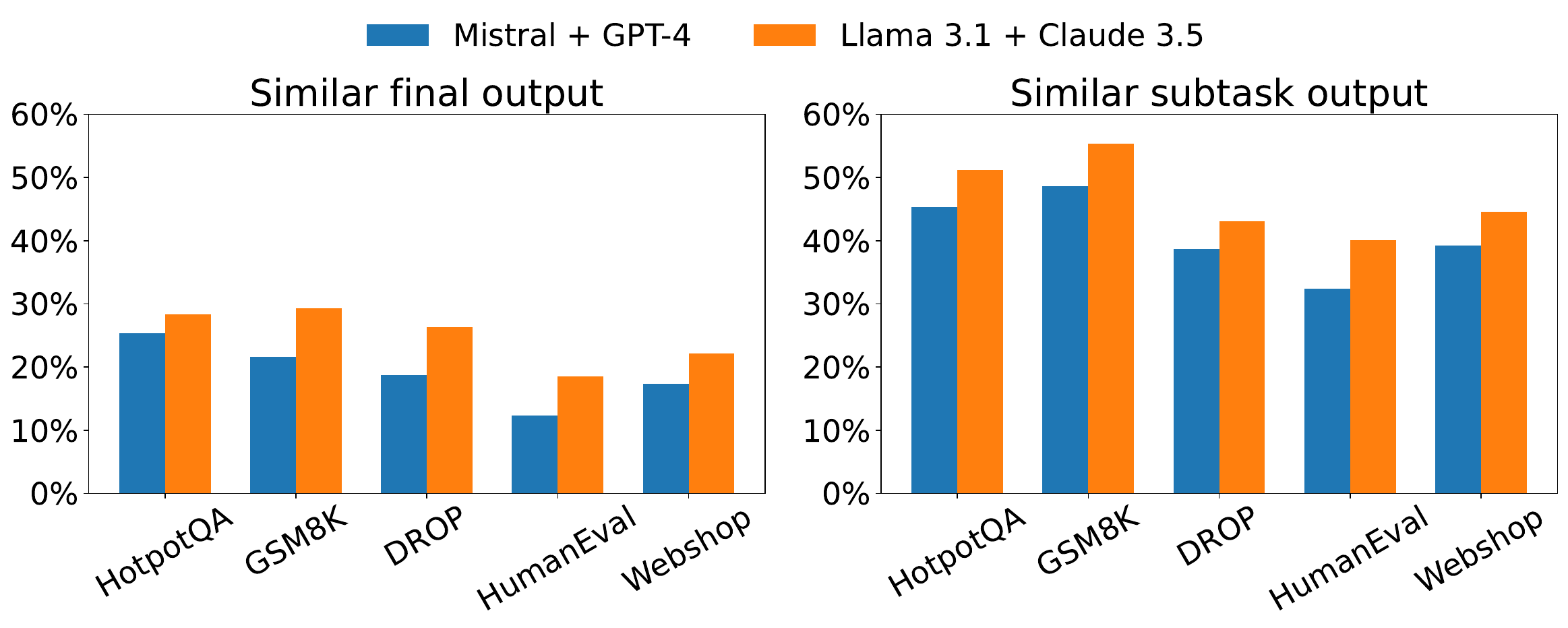}
    \caption{Similar output percentage across datasets.}
    \label{fig:output similarity}
\end{figure}

To further investigate subtask-level model assignment and identify which subtasks can be effectively handled by SLM, we conduct two additional experiments, presented in Figure \ref{fig:output similarity}: 1) the percentage of user requests where the final outputs are similar when processed entirely by either SLM or LLM (left), and 2) the average percentage of individual subtasks of a request processed by LLM, for which SLM produces similar subtask outputs. 
The figure demonstrates that a certain percentage (15.4\%-26.8\%) of user requests can be effectively managed by SLMs without sacrificing the accuracy of the final outputs. 
On average, 36.2\% to 51.9\% of subtask outputs for a request in LLMs are produced by SLMs. These findings indicate significant opportunities to reduce cloud usage by leveraging SLMs for suitable tasks and subtasks.

\begin{thm} 
\label{ob:3} The SLM can manage certain user requests and subtasks, producing outputs similar to the LLM.  (Figure \ref{fig:output similarity})
\end{thm}

\begin{wrapfigure}{r}{0.25\textwidth}
\includegraphics[width=0.24\textwidth, keepaspectratio]{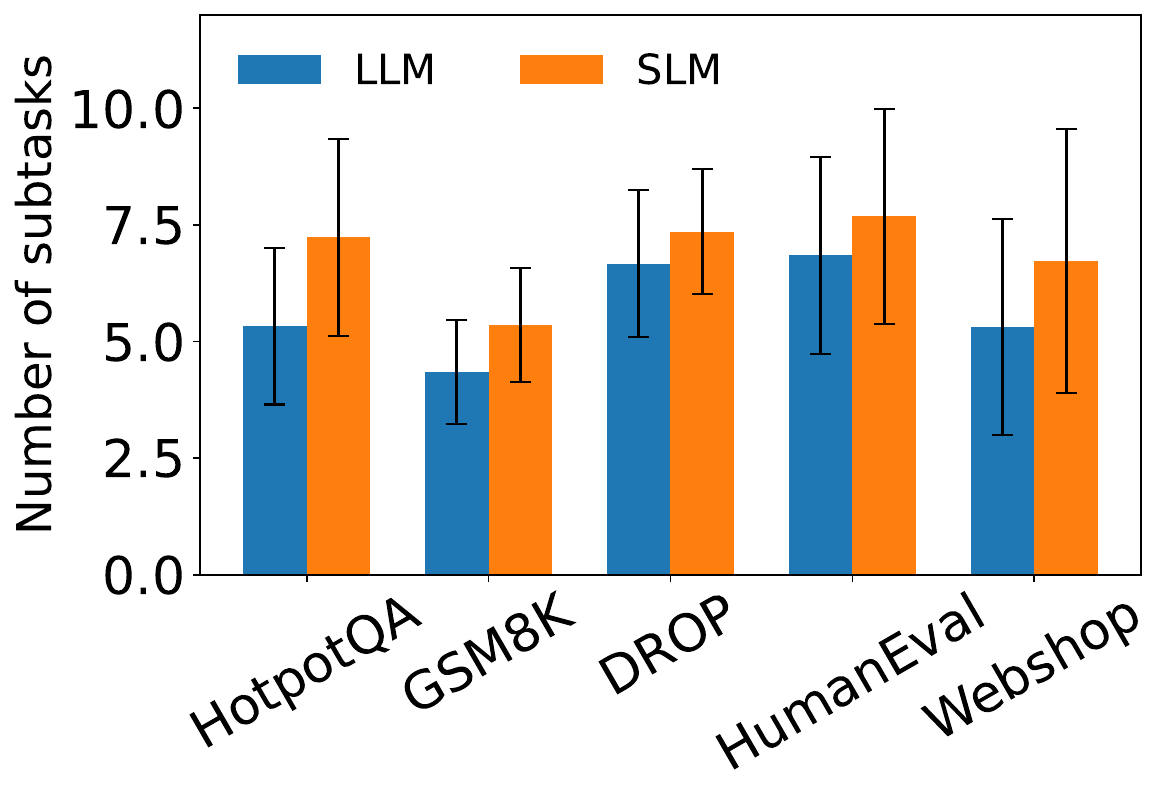}
\caption{Difference in number of subtasks in All-SLM vs. All-LLM.}
		\label{fig:num_sub}
\end{wrapfigure}

Figure \ref{fig:num_sub} illustrates the average number of subtasks generated per user request in All-SLM and All-LLM, respetively, across all datasets, considering only the requests that produced correct results. All-SLM generates more subtasks per request than All-LLM across all datasets, with SLM averaging 6.9 subtasks per user request compared to LLM's 5.8. This suggests that SLM breaks down requests into more granular subtasks, enabling a detailed, step-by-step approach. This decomposition arises from SLM's lower capability to handle complex requests, while LLM's superior ability allows it to generate fewer, more comprehensive subtasks.

\DEL{This suggests that SLM decomposes a request into more granular subtasks, potentially leading to a more detailed and step-by-step approach to task completion. This granular decomposition is likely due to SLM's lower capability in processing complex requests, necessitating smaller, more manageable steps. In contrast, LLM generates fewer subtasks due to LLM's higher capability in processing and understanding complex tasks as a whole.}

\begin{figure}
    \includegraphics[width=0.48\textwidth,height=\textheight,keepaspectratio]{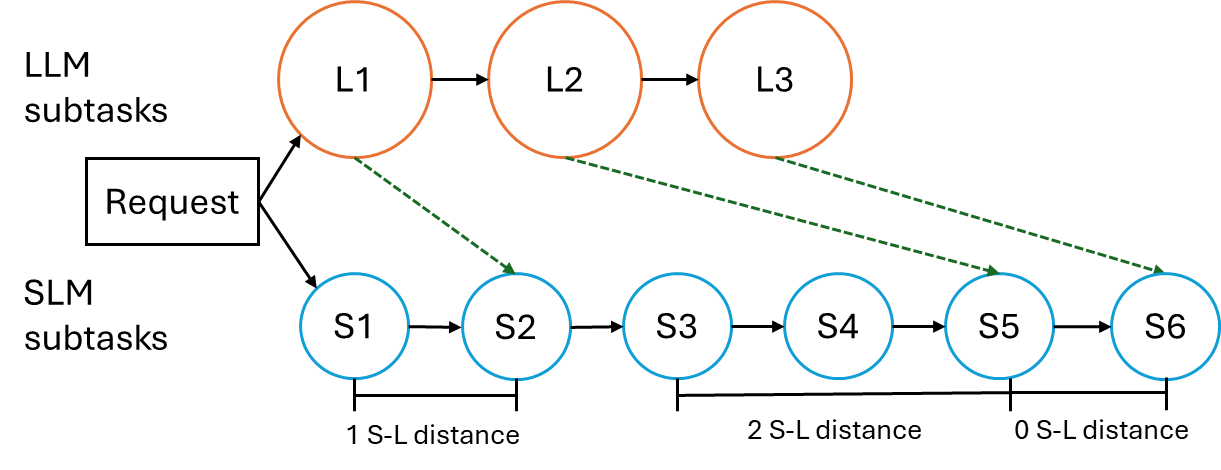}
    \caption{S-L distance illustration. }
    \label{fig:S-L}
\end{figure}

To evaluate the convergence of SLM and LLM subtasks, we introduce the concept of \emph{S-L distance} for an LLM subtask. This distance represents the number of additional SLM subtasks needed to produce a subtask similar (or match) to this LLM subtask, with their similarity defined as \emph{S-L similarity}. If no matched SLM subtask is found for an LLM subtask, its S-L distance is set to infinity. Figure \ref{fig:S-L} illustrates the S-L distance, where LLM-generated subtasks are denoted as L1, L2, and L3, while SLM-generated subtasks are denoted as S1, S2, S3, S4, S5, and S6. Dashed lines indicate matched outputs between SLM and LLM subtasks. Subtask L1's S-L distance is 1, indicating that one additional SLM subtask is needed to match it. L2 has an S-L distance of 2, requiring two extra SLM subtasks for a match. L3 directly corresponds to S6, resulting in an S-L distance of 0. This metric provides insight into the alignment between SLM and LLM outputs during request processing, highlighting how SLM subtask granularity compares to that of LLM at different stages.



\begin{figure}
    \centering
    \includegraphics[width=0.98\linewidth, keepaspectratio]{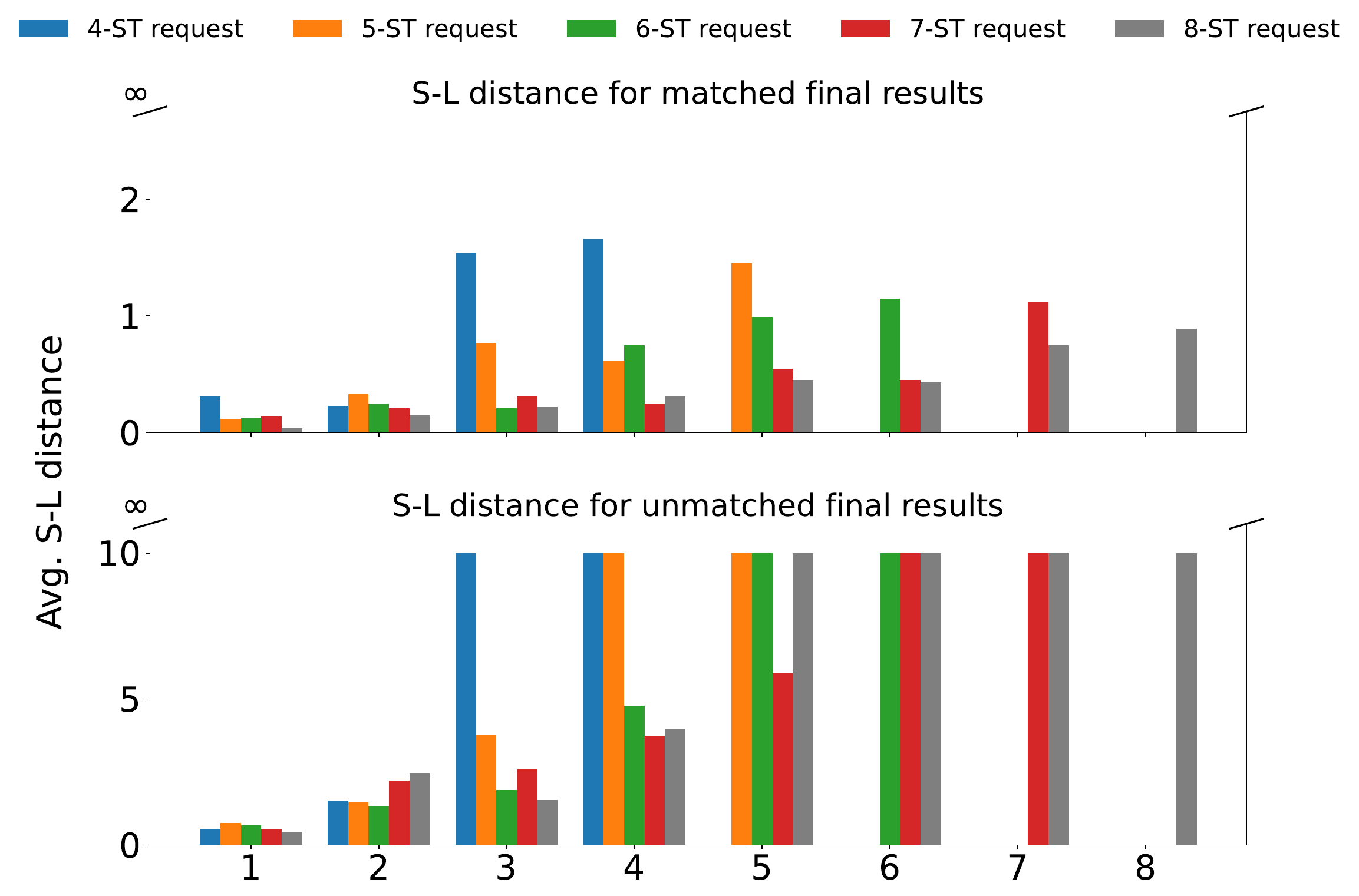}
    \caption{S-L distance comparison across subtask sequence.}
    \label{fig:s-l distance subtask}
\end{figure}

We ran each request using both All-LLM and All-SLM and categorized the requests into two groups: those with matched final results between LLM and SLM, and those without. Within each group, we further classified the requests based on the number of subtasks (ST) generated. Figure \ref{fig:s-l distance subtask} presents the average S-L distance across LLM subtasks with the same sequence ID for each request group for matched (top) and unmatched (bottom) final results scenarios. 
The request length means the number of subtasks for a request. As the subtask sequence progresses, the average S-L distance gradually increases in the matched scenario, while in the unmatched scenario, many distances reach infinity, indicating significant divergence between SLM and LLM outputs. This observation echoes Observation \ref{ob:2} that the later stage of subtasks is more important to the accuracy of the user request.

\begin{figure}
    \centering
    \includegraphics[width=0.98\linewidth, keepaspectratio]{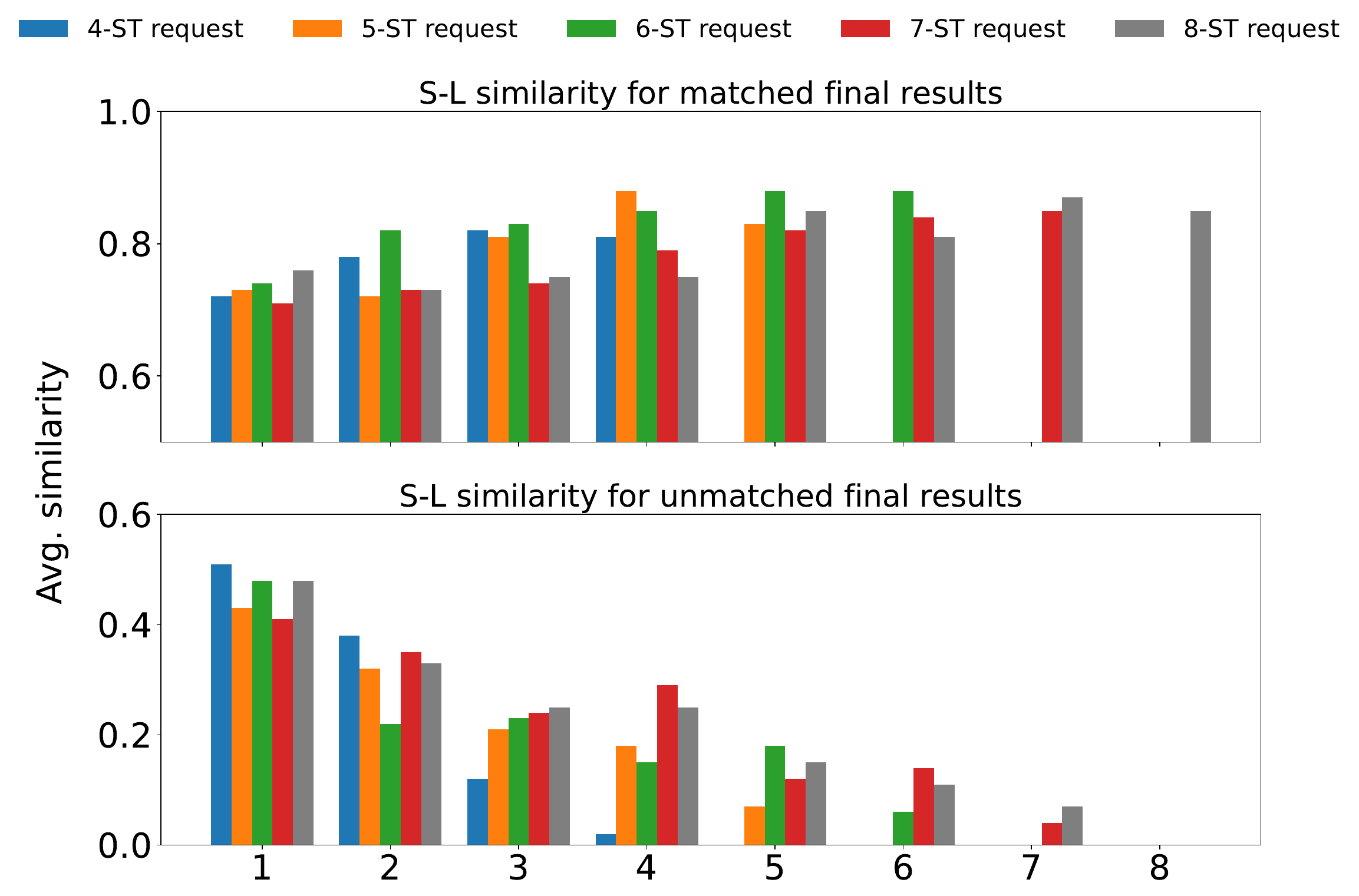}
    \caption{S-L similarity across subtask sequence IDs.} 
    \label{fig:s-l similarity}
\end{figure}

Figure \ref{fig:s-l similarity} shows the changes of S-L similarity as the request progresses in matched and unmatched scenarios. In matched cases, S-L similarity gradually increases, indicating that SLM outputs align more closely with LLM outputs in later stages. Conversely, in unmatched scenarios, S-L similarity remains low, suggesting persistent deviations. This measurement supports previous findings and highlights the potential for SLM and LLM convergence in later subtasks, enabling efficient subtask allocation. By starting with SLM and transitioning to LLM as similarity grows, we achieve LLM-level performance while minimizing cost.

\begin{thm} 
\label{ob:4} 
SLM typically generates more subtasks than LLM. Despite initial output deviations in the early stages, SLM may potentially converge with LLM's outputs in the later stages of request processing. (Figures \ref{fig:num_sub}- \ref{fig:s-l similarity})
\vspace{-0.0in}
\end{thm}

\section{Design of HERA}\label{sec3}

\subsection{Problem Statement and Overview}
\begin{figure*}
    \includegraphics[width=0.95\textwidth,keepaspectratio]{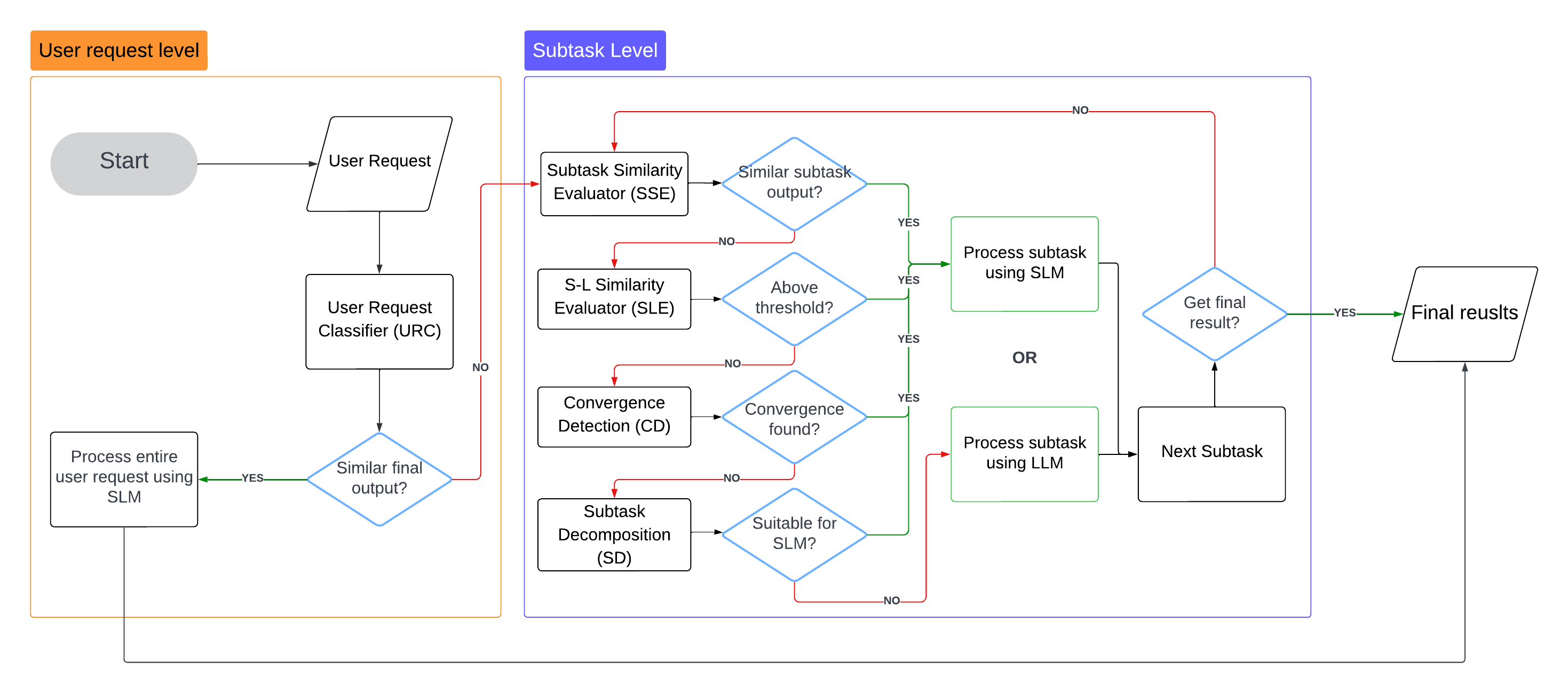}
    \caption{HERA decision-making workflow.}
    \label{fig:HERA}

\end{figure*}

Motivated by Observation~\ref{ob:1}, we propose HERA, which performs subtask allocation in the AI agent scenarios by dynamically assigning the subtasks of a request between the SLM and LLM in order to maximize the SLM usage while maintaining the LLM's accuracy of processing the request. Let \( R \) denote a given user request, and \( ST_i\) denote the $i^{th}$ dynamically generated subtask for request \( R \). The workflow of HERA is depicted in Figure \ref{fig:HERA}. First, based on Observation~\ref{ob:3}, HERA utilizes a lightweight classifier to determine if a user's request can yield similar outputs when executed entirely on either SLM or LLM. If yes, HERA opts for SLM for the entire request. 
Thus, HERA has a request-level classifier that determines whether the entire user request \( R \) can be processed by the SLM without compromising accuracy, denoted by \( c(R) = 0 \). 
If not, HERA proceeds to the next step, guided by Observation~\ref{ob:2}, to perform subtask-level model assignment. 
The objective of HERA is to design a \( router: ST_i \to \{0, 1\} \) such that each subtask \( ST_i \)  is routed to the SLM if \( router(ST_i) = 0 \), and to the LLM if \( router(ST_i) = 1 \). 
This hierarchical approach allows HERA to leverage the efficiency of SLM for user requests that can be accurately processed by SLM alone while enabling fine-grained subtask allocation for more complex requests.


In the subtask-level model assignment, HERA takes the following steps. 

\squishlist
\item [1)] Based on Observation~\ref{ob:3}, HERA estimates the output similarity of the $i^{th}$ subtask using LLM and SLM. If they are similar, HERA uses SLM for this subtask. Otherwise, HERA employs the following three steps based on Observation~\ref{ob:4}:

    \item [2)] It first estimates the S-L distance $d$ for the current subtask, followed by an evaluation of the S-L similarity. If the S-L similarity meets the required threshold (i.e., 0.7), the subtask is assigned to SLM, as it is expected to ultimately reach a similar subtask in LLM; otherwise, it proceeds to the next step.\looseness=-1
    
    \item [3)] It keeps estimating the output of the SLM subtask and LLM subtask and compares the results between each SLM and LLM subtask pair until it finds a convergence point where the S-L similarity reaches the threshold. 
    
    \item [4)] If no convergence point is identified, HERA breaks down the current subtask into smaller sub-subtasks to facilitate the SLM to process them. Only when all the decomposed sub-subtasks can be handled by the SLM, as determined in step 1), they will be processed by the SLM. Otherwise, the original subtask will be processed by the LLM.
    
\squishend

\DEL{Consequently, HERA provides an adaptive solution for model allocation in AI agent scenarios. It dynamically optimizes the collaboration between SLM and LLM, considering the unique characteristics of each user request and subtask, while leveraging the strengths of both models to achieve a balance between accuracy and cost-effectiveness.}

HERA consists of several offline-trained estimators based on profiled data. With the assistance from the estimators, after receiving a user request, HERA chooses between the SLM and LLM for the request or its subtask. 


\noindent \textbf{Offline Profiling.} HERA first profiles the AI agent that uses SLM and LLM with various user requests and their corresponding subtasks. The profiling process collects data on the subtask outputs from SLM and LLM. This data is then used to train prediction models, including the \emph{user request classifier}, \emph{subtask predictors} for the SLM and LLM execution, \emph{distance predictor}, and \emph{subtask decomposer}. These trained models are utilized in the online decision making of HERA to make informed decisions about task or subtask allocation.

\noindent \textbf{Online Decision Making.} After receiving a user request or a subtask, HERA determines its allocation between SLM and LLM. The decision-making process begins with the \emph{user request classifier}, which evaluates whether the entire request can be effectively handled using an SLM. If not, HERA proceeds to the subtask-level decision-making for a more granular analysis. At this stage, HERA employs \emph{subtask similarity evaluator}, \emph{S-L similarity estimator} (SLE), \emph{convergence detector} (CD), and \emph{subtask decomposer} (SD) approaches to determine the suitable model for processing the current subtask. HERA makes decisions for every subtask using this process until the final result is output.

\textcolor{black}{Note that this paper is for initial exploration of hybrid AI agent deployment, we focus on optimizing the primary cost driver - LLM API usage - while maintaining accuracy. While real-world deployments involve additional considerations like edge computing costs, network bandwidth constraints, and SLA requirements, our results demonstrate that even this simplified model can achieve significant cost reductions. The framework can be extended to incorporate more complex cost models as discussed in Section \ref{sec: 5}.}

\subsection{Offline Profiling}
In the offline profiling phase, HERA collects data to fine-tune models that predict the performance of different subtask allocations between SLM and LLM. The profiling is conducted on user requests from historical trace datasets (e.g., GSM8K and HotPotQA) and their corresponding subtasks.

\noindent\textbf{Data Collection:} For each user request \( R \), we generate a binary tree of subtasks, where each node represents a subtask \( ST_i \) and each edge represents using a model (SLM or LLM) to process the parent subtask. Starting from the root node, which represents the initial user request, we process the subtask using both the SLM and LLM, creating two child nodes. For each child node, we then recursively process the corresponding subtask using both SLM and LLM, further creating child nodes until a predefined depth (e.g., 15 subtasks) is reached or the model thinks the request is finished. At each leaf node, we profile the output of executing the subtask using the selected model. 
In addition to the subtask-level profiling, we also profile the performance of executing the entire user request \( R \) using SLM and LLM. The similarity of the final results from the two models is collected. Moreover, we use the SLM to generate multiple smaller subtasks for each original subtask, creating a dataset of subtask decomposition. Using the collected profiled data, we train the following models.





\noindent \textbf{User Request Classifier (URC)}: 
For each user request in the profiled data, we use the user request as the input feature and the similarity score between the outputs generated by processing the entire user request using All-SLM and All-LLM as the target variable. We then train the \emph{user request classifier} model using this input-output data. 



\noindent \textbf{Subtask Predictor (SP)}: We train two separate models, $SP_{SLM}$ and $SP_{LLM}$, which learn to predict the next subtask when the current subtask is processed using SLM and LLM, respectively. For each node in the binary tree of subtasks generated during the data collection process, we use the subtask at that node as the input feature and the subtask generated by applying SLM or LLM to the current subtask as the target output for the two models, respectively. 

\noindent \textbf{Distance Predictor (DP)}: The distance predictor predicts the S-L distance. For each request in the profiled data, we extract the content of the LLM subtask and its sequence ID as the input features and its corresponding S-L distance as the target label. We then train the distance predictor model using the input and output data to predict the S-L distance.

\noindent \textbf{Subtask Decomposer (SD)}: The Subtask Decomposer is trained to break down a complex subtask into smaller, more manageable sub-subtasks. It takes a subtask and the predicted next subtask from $SP_{LLM}$ for this subtask as inputs, and outputs a sequence of sub-subtasks, aiming to ensure that the output of the last sub-subtask is similar to the predicted next subtask from $SP_{LLM}$. The \emph{subtask decomposer} is trained using data derived from decomposing subtasks from user request traces.

For the above models, we use DeBERTa~\cite{he2021debertav3} as the base for URC and DP, and Llama 3.2 1B~\cite{llama3} for SP and SD, fine-tuning them with LoRA~\cite{hu2021lora}. These models are then used in the subsequent components of HERA to make informed decisions about task or subtask allocation. 

\subsection{Online Decision Making}\label{sec3.3}
\subsubsection{General Process}
Algorithm \ref{alg:HERA} shows the pseudocode of the decision-making process of HERA. When a user request is received, the \emph{user request classifier} (URC) first predicts the similarity score of running the entire request on SLM and LLM. If the similarity score exceeds a threshold, the request is processed entirely by SLM. Otherwise, it advances to subtask-level decision-making. Specifically, the \emph{subtask similarity evaluator} (SSE) compares the predicted outputs of the subtask from SLM and LLM using $SP_{SLM}$ and $SP_{LLM}$. If the outputs are similar, the subtask is assigned to SLM. 
If not, the \emph{S-L similarity estimator} (SLE) component, using the \emph{distance predictor} (DP), estimates the S-L distance of the current subtask ($d$). \textcolor{black}{It then uses $SP_{SLM}$ to predict the next subtask consecutively for $d+1$ times to generate a sequence of $d+1$ future SLM subtasks, and uses $SP_{LLM}$ once to predict the next LLM subtask. Next, SLE compares the  $(d+1)^{th}$ SLM subtask with the predicted LLM subtask to determine if they match. If yes, SLM is used; otherwise, the process moves to \emph{convergence detector} (CD). It attempts to identify a convergence point, where the outputs of the SLM and LLM are similar. If it is found, SLM is used until the convergence point. If not, the \emph{subtask decomposer} (SD) breaks the subtask into smaller sub-subtasks, and the process of the \emph{SSE} repeats for each sub-subtask. }

\subsubsection{Request-Level Decision Making}


HERA firstly uses the \emph{user request classifier} to process an incoming user request. It leverages the knowledge learned during the offline profiling phase to identify user requests that can be accurately processed by the SLM alone, avoiding unnecessary subtask-level allocation. Specifically, HERA feeds the request into the \emph{user request classifier} model. The classifier predicts a similarity score between 0 and 1, indicating the expected similarity between the results of processing the request solely using SLM versus the LLM. If the predicted similarity score is above a predefined threshold (e.g., 0.7), HERA processes the entire user request using SLM, bypassing the subtask-level allocation. Otherwise, LLM is used to produce a subtask and HERA proceeds to the subtask-level allocation. 

\begin{algorithm}
\caption{HERA online decision making process.}
\label{alg:HERA}
\begin{algorithmic}[1]
\Require User request R
\Ensure Final result
\If{URC(R) predicts similar output}
    \State Process R using SLM
\Else
    \For{each subtask $ST_i$ generated}
        \If{SSE($ST_i$) predicts similar outputs}
            \State Process $ST_i$ using SLM
        \ElsIf{SLE($ST_i$) finds high S-L similarity}
            \State Process $ST_i$ and next d-1 subtasks using SLM
        \ElsIf{CD($ST_i$) finds convergence point}
            \State Process subtasks to convergence using SLM
        \Else
            \State sub\_subtasks = SD($ST_i$)
            \For{each sub\_subtask}
                \State Recursively apply SSE
            \EndFor
        \EndIf
    \EndFor
\EndIf
\State \Return Final result
\end{algorithmic}
\end{algorithm}

\subsubsection{Subtask-Level Decision Making}
If the \emph{user request classifier} determines that a user request requires subtask-level allocation, HERA proceeds to the SSE to process the subtask. 

\noindent \textbf{Subtask Similarity Evaluator (SSE)}: The \emph{subtask similarity evaluator} compares the outputs of the SLM and the LLM for each subtask, assessing their similarity and making appropriate model assignments based on the stage of the user request. For each subtask $ST_i$ in the user request $R$, HERA feeds the current subtask into the $SP_{SLM}$ and $SP_{LLM}$ models. The $SP_{SLM}$ and $SP_{LLM}$ models generate the predicted next subtasks for the SLM and the LLM, respectively. 
The \emph{subtask similarity evaluator} then estimates the similarity of the predicted next subtasks as introduced in Section \ref{sec:2.1}. If the similarity is above a predefined threshold $\kappa$, the subtask $ST_i$ is assigned to the SLM. The similarity threshold $\kappa$ is determined through empirical analysis during the offline profiling phase. For each SLM-LLM pair, we analyze a set of requests and their subtasks, measuring the relationship between threshold values and final accuracy. The detailed sensitivity analysis of the threshold can be found in Section \ref{sec: 4.6}. Organizations deploying HERA can fine-tune these thresholds during their offline profiling phase based on their specific accuracy requirements and cost constraints.

Here, instead of using a constant similarity threshold, based on Observation \ref{ob:2}, we set the threshold adaptively based on the subtask's sequence ID. The threshold $\kappa$ is smaller at the early stages of a request, permitting loose comparisons, and increases as the request progresses, making it more stringent in the later stages. Guided by Observation \ref{ob:2}, we let $\kappa$ increase linearly with the subtask's sequence ID, calculated as $\kappa = threshold_{base}+\min(ID, 5) * 0.02$, where $threshold_{base} = 0.6$ and all these parameters are determined empirically. 
 
If the similarity is below the threshold $\kappa$, the simplest way is to use LLM. However, directly using LLM results in lower SLM usage. To address this problem, we employ three approaches: the \emph{S-L similarity evaluator}, the \emph{convergence detector}, and the \emph{subtask decomposer}. The S-L similarity evaluator identifies when a future SLM subtask matches the current LLM subtask, the convergence detector finds matching future subtasks between the SLM and LLM, and the subtask decomposer breaks down the current subtask into smaller subtasks to increase the likelihood of processing by the SLM. The details are presented in the following.


\noindent \textbf{S-L Similarity Evaluator (SLE)}:
Guided by Observation \ref{ob:4}, the S-L distance metric helps determine if a future SLM subtask matches the current LLM subtask, which is crucial for deciding whether to process a subtask using SLM or LLM. Thus, the S-L similarity evaluator dynamically adjusts the similarity threshold during task processing based on the progress stage of the user request. It receives the current subtask $ST_i$ and its sequence ID as inputs and uses the \emph{distance predictor} model to estimate the S-D distance $d$ between the outputs of the SLM and the LLM, considering the current subtask's content. The $SP_{SLM}$ model then predicts the output for the $(i+d)^{th}$ subtask, while the $SP_{LLM}$ predicts the output for the $i^{th}$ subtask. These outputs are compared using the predefined similarity threshold $\kappa$ (same threshold as in \emph{subtask similarity evaluator}), and the subtask $ST_i$ is assigned to the SLM; otherwise, we proceed to the next component. 




\noindent \textbf{Convergence Detector (CD)}: Guided by Observations \ref{ob:2} and \ref{ob:4}, \emph{convergence detector} identifies a future convergence point between the outputs of SLM and LLM. Starting from subtask $ST_i$, the \emph{convergence detector} uses $SP_{SLM}$ and $SP_{LLM}$ to predict future subtasks iteratively. It compares the similarity of each pair of the SLM and LLM predictions using the same similarity metric and threshold as previous components. It continues this process for a predefined number of future subtasks or until the end of the sequence. If multiple convergence points are found, \emph{convergence detector} selects the latter one to increase the use of SLM. All subtasks from $ST_i$ up to the identified convergence point are then assigned to SLM. This approach allows HERA to maximize SLM usage when eventual alignment with LLM outputs is predicted, even if initial subtasks diverge. If no convergence is detected, we proceed to the next component.

\noindent \textbf{Subtask Decomposer (SD)}: 
Guided by Observation \ref{ob:4}, which highlights the granularity and step-by-step nature of SLM processing, we design the \emph{subtask decomposer}. It breaks down a complex subtask into smaller sub-subtasks, making them easier for the SLM to process. It takes a current subtask $ST_i$ as input and uses the \emph{subtask decomposer} model, which is trained during the offline profiling phase, to generate a sequence of sub-subtasks, denoted by $\{SST_1, SST_2, \ldots, SST_m\}$. HERA then evaluates each sub-subtask $SST_j$ to determine its suitability for processing by the SLM. Specifically, HERA inputs $SST_j$'s content into both $SP_{SLM}$ and $SP_{LLM}$ models, which then predict the next sub-subtask. If the similarity of the two predicted next sub-subtasks exceeds the predefined threshold $\kappa$, the sub-subtask is deemed suitable for SLM processing. Only when all sub-subtasks are found suitable for the SLM, HERA assigns all sub-subtasks $\{SST_1, SST_2, \ldots, SST_m\}$ to SLM to produce the output for the original subtask. Conversely, if any sub-subtask is unsuitable for SLM, HERA assigns the subtask $ST_i$ to the LLM. While we could allocate each sub-subtask individually to the SLM or LLM, this may increase the number of LLM calls. To avoid this, we allocate the entire group of decomposed sub-subtasks or the original subtask as a single unit. 





\section{Performance Evaluation}\label{sec5}






\subsection{Experiment Settings} 

\DEL{\noindent\textbf{Models, environments, and baselines.} \textcolor{black}{The experimental settings, including the primary model pairs used (Mistral 7B+GPT-4 and Llama 3.1 8B+Claude sonnet 3.5) and software and hardware environment, are the same as those described in Section \ref{sec:2.1}, unless otherwise specified. We compare HERA against several baseline methods: SLM, LLM, Random assignment, and HybridLLM. HybridLLM was implemented and trained using the same dataset as HERA for fair comparison.} \sh{remove anything that is the same as the Data Analysis section. Just say the settings are the same as those in Section xx unless otherwise specified. The same for the following 2 paragraphs.-done}}

The experiment settings are the same as those in Section~\ref{sec:2.1} unless otherwise specified. 
We employed GPT-4+Mistral 7B as the LLM-SLM pair to generate 1000 subtask traces using HotpotQA \cite{yang2018hotpotqa} and GSM8K \cite{cobbe2021training} to fine-tune the models in HERA. 
The fine-tuning process for all estimators took approximately 2 hours on a cloud-based Nvidia A100, a one-time cost that enables subsequent efficient decision-making. To evaluate the performance and generalization capabilities of HERA, we test on six benchmarks: HotpotQA, GSM8K, DROP, HumanEval, Webshop and MATH \cite{hendrycksmath2021} though we trained the HERA in only two of the datasets.  



\subsection{Overall Performance}



\begin{figure}
    \includegraphics[width=0.48\textwidth,height=\textheight,keepaspectratio]{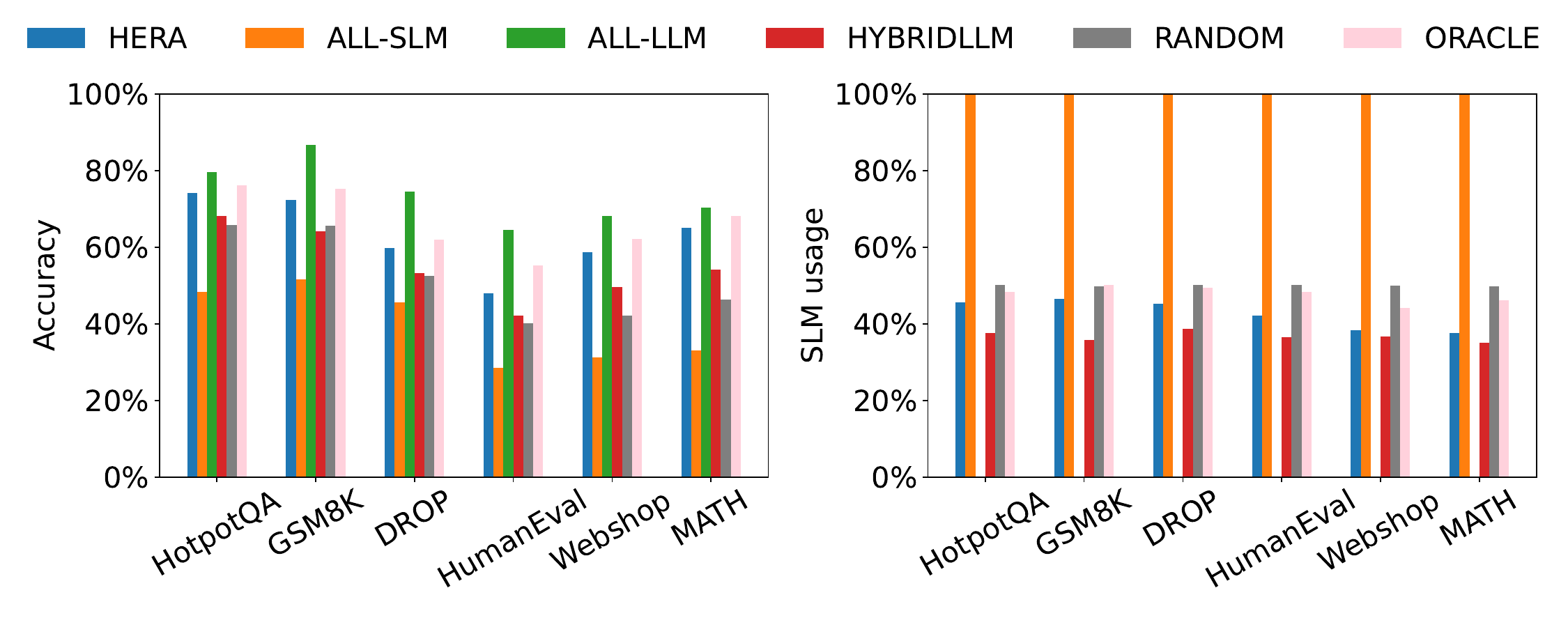}
   \caption{Accuracy and SLM usage.}
    \label{fig:acc_slm}

\end{figure}

\noindent \textbf{Accuracy and SLM usage.} 
Figure \ref{fig:acc_slm} shows the accuracy and SLM usage comparison across six datasets for different methods. HERA demonstrates higher performance across all datasets, balancing high accuracy with efficient SLM usage. Starting with the datasets that are used for training HERA. For HotpotQA, HERA achieves 74.12\% accuracy (vs. 79.67\% for LLM, 76.25\% for Oracle) while using SLM for 45.67\% of subtasks (vs.48.34\% for Oracle). For GSM8K, HERA attains 72.34\% accuracy, surpassing SLM (51.56\%) and HybridLLM (64.23\%), with 46.45\% SLM usage. In other datasets, including DROP, HumanEval, MATH and Webshop, HERA consistently delivers high accuracy (59.78\%, 47.89\%, 65.3\% and 58.7\%), outperforming HybridLLM by 6.56\%, 5.67\%, 10.88\% and 9.10\%, respectively, while maintaining SLM usage between 37.65\% and 45.34\%. The performance gap between HERA and LLM is larger for DROP, HumanEval, and Webshop compared to HotpotQA and GSM8K since HERA is fine-tuned on the latter two. Despite this, HERA demonstrates robustness across diverse datasets, effectively reducing reliance on costly LLM computations while maintaining competitive accuracy.

\DEL{Stacked bars with different patterns illustrate the contribution of SLM latency, LLM latency, and HERA overhead to the total latency for each method. From the results, }


\noindent\textbf{Call-Performance Threshold (CPT)}. Based on \cite{ong2024routellm}, we used CPT(x\%) to represent the minimum percentage of LLM calls needed to achieve x\% of the accuracy gap between All-SLM and All-LLM. Lower CPT values indicate fewer expensive LLM calls are needed to achieve a high accuracy, or better system effectiveness. Table \ref{tab:cpt-values} shows the CPT values for HERA, HybridLLM, and Random across the six datasets. To calculate the CPT values, we first determined the accuracy gap between All-SLM and All-LLM for each dataset. For each method (HERA, HybridLLM, and Random), we varied their respective decision-making parameters across a reasonable range. Specifically, for HERA and HybridLLM, we adjusted the similarity threshold, and for Random, we varied the probability of choosing the LLM. For each parameter configuration, we recorded the corresponding percentage of LLM calls and the resulting accuracy. These results were then sorted by the percentage of LLM calls, generating an accuracy-versus-LLM-usage curve. From this curve, we identified the minimum percentage of LLM calls needed to achieve 50\%, 70\%, and 90\% of the accuracy gap as CPT(50\%), CPT(70\%), and CPT(90\%), respectively. 

HERA consistently outperforms HybridLLM and Random allocation across all datasets at each CPT level. At CPT(50\%), HERA shows the most significant improvements. For instance, in HotpotQA, HERA requires only 38.77\% LLM calls compared to HybridLLM's 54.33\% and Random's 59.88\%. As CPT level increases, the gap between HERA and other methods narrows, but HERA maintains its advantage. This can be attributed to the increasing complexity of latter subtasks, which necessitates LLM usage for all systems. HERA's persistent advantage, even at CPT(90\%), demonstrates its capability to identify subtle opportunities for SLM usage without compromising accuracy performance across diverse tasks.

\begin{table}[h]
\centering
\caption{CPT values at different thresholds across datasets (the lower the better).}
\label{tab:cpt-values}
\small
\setlength{\tabcolsep}{4pt}  
\begin{tabular}{@{}l@{\hspace{18pt}}l@{\hspace{4pt}}c@{\hspace{4pt}}c@{\hspace{4pt}}c@{}}
\hline
Dataset & Method & CPT(50\%) & CPT(70\%) & CPT(90\%) \\
\hline
\multirow{3}{*}{\makebox[0.8cm][l]{HotpotQA}}
 & HERA        & 38.8 & 55.3 & 81.6 \\
 & HybridLLM   & 54.3 & 62.5 & 83.8 \\
 & Random      & 59.9 & 78.8 & 95.7 \\
\hline
\multirow{3}{*}{\makebox[0.8cm][l]{GSM8K}}
 & HERA        & 46.7 & 59.2 & 84.6 \\
 & HybridLLM   & 57.7 & 65.3 & 84.6 \\
 & Random      & 54.1 & 77.2 & 96.1 \\
\hline
\multirow{3}{*}{\makebox[0.8cm][l]{DROP}}
 & HERA        & 40.3 & 57.9 & 82.5 \\
 & HybridLLM   & 55.4 & 63.7 & 84.2 \\
 & Random      & 59.8 & 78.9 & 96.0 \\
\hline
\multirow{3}{*}{\makebox[0.8cm][l]{HumanEval}}
 & HERA        & 44.3 & 62.5 & 86.8 \\
 & HybridLLM   & 59.8 & 67.9 & 85.7 \\
 & Random      & 56.9 & 77.0 & 92.2 \\
\hline
\multirow{3}{*}{\makebox[0.8cm][l]{Webshop}}
 & HERA        & 44.8 & 60.5 & 88.3 \\
 & HybridLLM   & 63.3 & 75.4 & 91.2 \\
 & Random      & 65.2 & 77.2 & 91.4 \\
\hline
\multirow{3}{*}{\makebox[0.8cm][l]{MATH}}
 & HERA        & 46.1 & 62.2 & 88.1 \\
 & HybridLLM   & 68.4 & 77.6 & 93.6 \\
 & Random      & 68.1 & 82.1 & 95.2 \\
\hline
\end{tabular}
\end{table}

\begin{figure}
    \includegraphics[width=0.48\textwidth,height=\textheight,keepaspectratio]{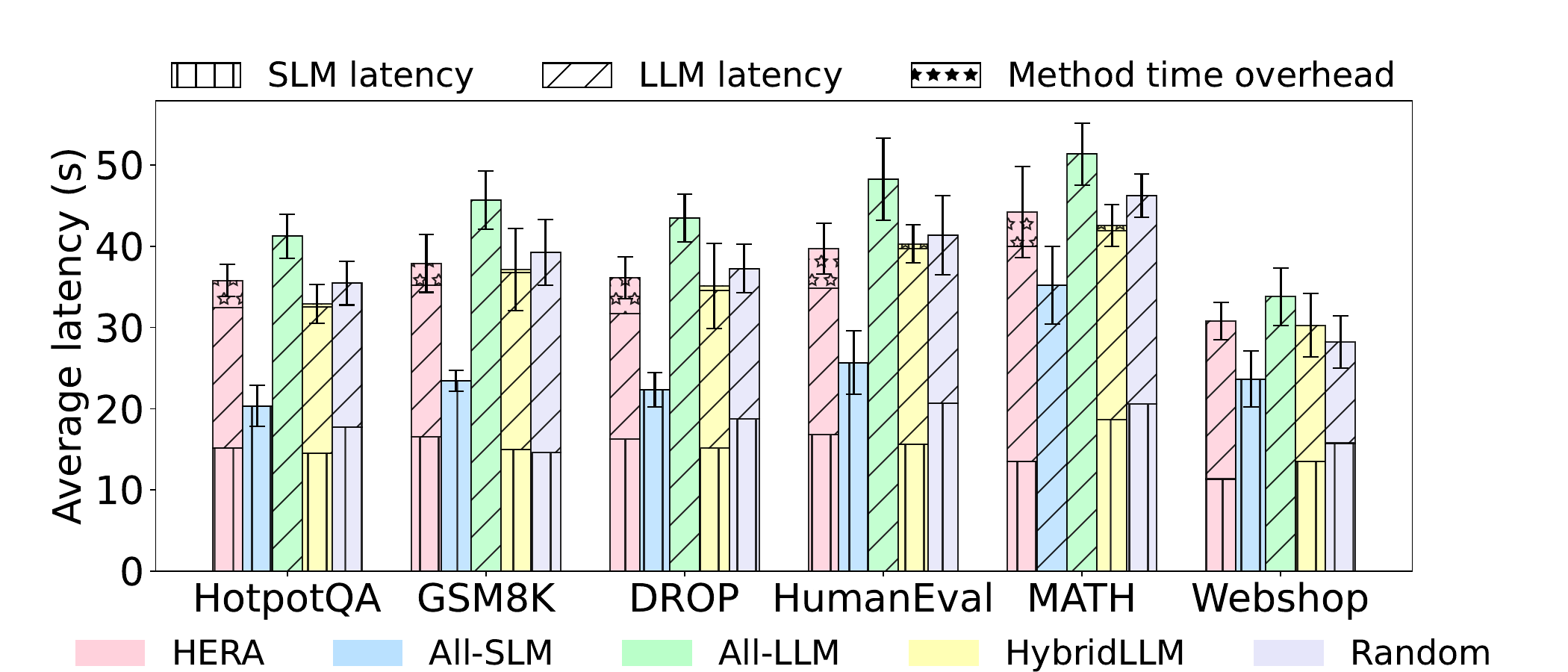}
    \caption{Latency breakdown.} 
    \label{fig:latency}

\end{figure}

\noindent\textbf{Average latency.}
We measured the average latency to show that HERA does not compromise the latency though it is not a focus in this paper. Figure \ref{fig:latency} shows the average latency breakdown across different methods and datasets. The average latency is decomposed to the SLM latency, LLM latency and the method time overhead. HERA achieves latency lower than All-LLM, higher than All-SLM, and comparable to HybridLLM and Random across all datasets. For instance, in HotpotQA, HERA has a total latency of 35.78s, lower than All-LLM (41.23s), higher than All-SLM (20.34s), and similar to HybridLLM (32.56s) and Random (35.45s). This trend is consistent across all datasets. The SLM produces lower latency than the LLM by eliminating the need to communicate with distant cloud servers and relying on a smaller model. 
By moving some subtasks to SLM, HERA reduces the latency of All-LLM. The HERA time overhead, representing additional decision-making time, averages 2.73s to 4.87s across datasets, constituting 8-10\% of total HERA latency. This overhead remains relatively minor compared to overall latency and is offset by reduced LLM reliance. The time overhead of HybridLLM (0.26s) is negligible compared to the typical request processing times of 30-40s. We found that the average network latency for transmitting a subtask from the cloud-based LLM to the local SLM and receiving its response, excluding processing time, is 0.58 seconds. This latency is negligible compared to the overall average latency. In summary, despite utilizing significantly less powerful local hardware compared to cloud infrastructure, HERA achieves comparable or better latency than the cloud-only approach. This suggests that offloading some subtasks to the local edge device will not incur high communication latency.

\DEL{\textcolor{black}{The average network latency for transmitting requests and receiving responses from the cloud-based LLM (0.58s) and the computational overhead of HybridLLM (0.26s) is negligible compared to the typical request processing times of 30-40s. This network latency includes the time for sending the request to the cloud server and receiving the response, while HybridLLM overhead represents the additional time required for its decision-making process.}}


\DEL{To evaluate the economic efficiency of HERA, we conducted a cost analysis across different methods and datasets. }

\noindent\textbf{Monetary cost.}
The monetary cost is primarily based on the usage of the LLM (GPT-4), while the use of the SLM (Mistral 7B) is considered free due to its open-source nature and local deployment. In this analysis, we focused solely on computational costs and did not consider the data transfer costs between cloud and the edge device. Table \ref{tab:total-cost-comparison} presents the average USD per request for processing each dataset using different methods. HERA consistently reduces the cost by 19-30\% compared to All-LLM across all datasets. This substantial reduction is achieved through the intelligent allocation of subtasks to the free SLM when appropriate, without significantly compromising accuracy. 
\textcolor{black}{For perspective, in a production environment processing 1 million requests monthly, this translates to potential savings of \$9,000-\$26,000 per month (based on current GPT-4o pricing) with minimal impact on service quality. These savings become even more significant for organizations handling larger request volumes.}
In summary, HERA offers a cost-effective solution for deploying AI agents, particularly when balancing performance and operational costs is crucial.


\begin{table}[h]
\centering
\caption{Cost (\$/request) comparison.}
\label{tab:total-cost-comparison}
\begin{tabular}{lccccc}
\hline
Dataset & HERA & All-LLM & HybridLLM & Random \\
\hline
HotpotQA & 0.043 & 0.054 & 0.046 & 0.027 \\
GSM8K & 0.035 & 0.043 & 0.037 & 0.031 \\
DROP & 0.034 & 0.044 & 0.039 & 0.029 \\
HumanEval & 0.044 & 0.056 & 0.046 & 0.039 \\
Webshop & 0.058 & 0.084 & 0.067 & 0.048 \\
MATH & 0.033 & 0.044 & 0.041 & 0.024 \\
\hline
\end{tabular}
\end{table}

\textcolor{black}{In summary, HERA achieves a strategic balance between efficiency and performance. While it shows a 4-6\% accuracy reduction compared to All-LLM, this enables 19-30\% cost savings (\$2,100-\$5,200 monthly for 100,000 requests) with comparable latency. These benefits persist even as LLM costs evolve, offering consistent value through reduced network dependency and adjustable thresholds. For production environments, the substantial cost advantages typically justify the minimal accuracy trade-off, particularly since affected requests remain within acceptable performance parameters.}

\subsection{Generalizability Evaluation}


Table \ref{tab:generalizability} shows HERA's performance across two model pairs: Mistral 7B+GPT-4 and Llama-3 8B+Claude 3.5. ($\pm$x\%) in the table indicates that the performance metric changes by x\% when switching from Mistral+GPT-4 to Llama+Claude.
HERA demonstrates robust performance across both model pairs, with accuracy differences ranging from -3.66\% to +4.33\% when switching model pairs. For instance, HotpotQA shows an accuracy increase of 4.33\% with Llama-3+Claude, while DROP experiences a 3.55\% decrease. SLM usage varies between -4.65\% to +8.20\%, indicating HERA's ability to generalize its routing strategy to each model pair's characteristics. Notably, cost changes vary significantly across datasets, from a 29.31\% decrease for Webshop to a 20.45\% increase for HumanEval. This variation suggests that the cost-effectiveness depends on the specific task and model pair combination. These results demonstrate HERA's generalization capabilities. Despite slight performance variations between model pairs, HERA maintains its efficiency in balancing SLM and LLM usage across diverse datasets without retraining, indicating its potential adaptability to different applications and data distributions in practice.


\begin{table}[h]
\centering
\caption{Generalizability of HERA for different model pairs.}
\label{tab:generalizability}
\small
\begin{tabularx}{\columnwidth}{@{}lXcc@{}}
\toprule
Dataset & Metric & Mistral+GPT-4 & Llama-3+Claude \\
\midrule
\multirow{3}{*}{\rotatebox[origin=c]{0}{HotpotQA}} 
& Acc. (\%)        & 74.1 & 78.4 (+4.30\%) \\
& SLM (\%)  & 45.6 & 48.8 (+3.22\%) \\
& Avg. cost (\$) & 0.043 & 0.036 (-16.28\%) \\
\midrule
\multirow{3}{*}{\rotatebox[origin=c]{0}{GSM8K}} 
& Acc. (\%)        & 72.3 & 73.7 (+1.40\%) \\
& SLM \%  & 46.4 & 48.1 (+1.67\%) \\
& Avg. cost (\$) & 0.035 & 0.033 (-5.71\%) \\
\midrule
\multirow{3}{*}{\rotatebox[origin=c]{0}{DROP}} 
& Acc.(\%)        & 59.7 & 56.2 (-3.50\%) \\
& SLM \% & 45.3 & 42.6 (-2.67\%) \\
& Avg. cost (\$) & 0.034 & 0.038 (+11.76\%) \\
\midrule
\multirow{3}{*}{\rotatebox[origin=c]{0}{HumanEval}} 
& Acc. (\%)        & 47.8 & 44.2 (-3.60\%) \\
& SLM \%  & 28.4 & 36.6 (+8.20\%) \\
& Avg. cost (\$) & 0.044 & 0.053 (+20.45\%) \\
\midrule
\multirow{3}{*}{\rotatebox[origin=c]{0}{Webshop}} 
& Acc. (\%)        & 58.7 & 61.3 (+2.60\%) \\
& SLM \%  & 38.3 & 33.6 (-4.65\%) \\
& Avg. cost (\$) & 0.058 & 0.041 (-29.31\%) \\
\midrule
\multirow{3}{*}{\rotatebox[origin=c]{0}{MATH}} 
& Acc. (\%)        & 65.4 & 72.3 (+6.90\%) \\
& SLM \%  & 37.65 & 41.35 (+3.70\%) \\
& Avg. cost (\$) & 0.033 & 0.036 (+9.09\%) \\
\bottomrule
\end{tabularx}
\end{table}

\subsection{Ablation Study}

\begin{figure}
    \centering
    \includegraphics[width=0.45\textwidth,height=\textheight,keepaspectratio]{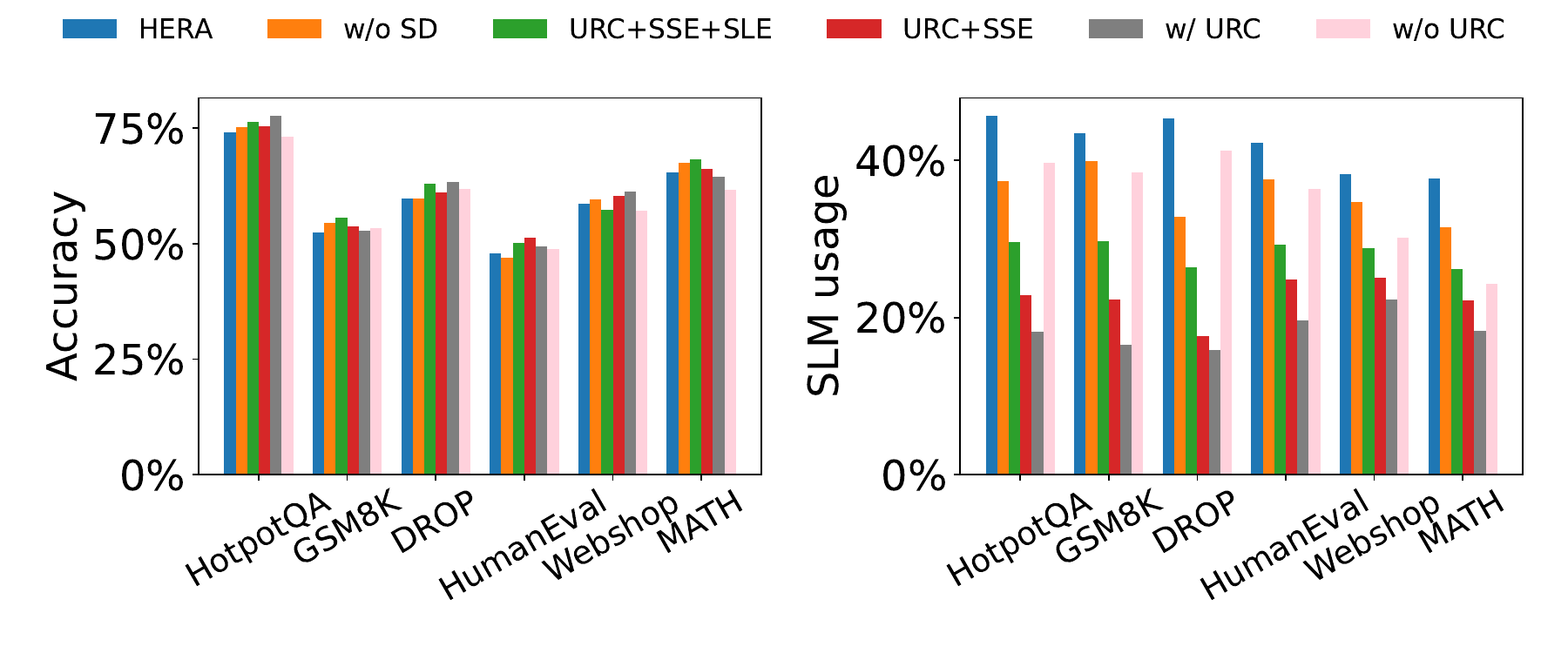}
    \vspace{-0.2in}
    \caption{Ablation study.}
     \vspace{-0.1in}
    \label{fig:Ablation}
\end{figure}

Figure \ref{fig:Ablation} shows the accuracy and SLM usage of HERA after we gradually remove individual components. 
Recall that HERA incorporates the \emph{user request classifier} (URC), \emph{subtask similarity evaluator} (SSE), \emph{S-L similarity evaluator} (SLE), \emph{convergence detector} (CD), and the \emph{subtask decomposer} (SD). We use `w/o SD' to denote HERA without the SD component, `URC+SSE+SLE' for the combination of these three components, `w/URC' for HERA with only URC, and `w/o URC' for HERA without URC. We observe that removing components often leads to increases in accuracy but at the cost of reduced SLM usage. 
For instance, removing the \emph{subtask decomposer} (w/o SD) results in minor accuracy increases for most datasets (0.08\%-2.14\%) but reduces SLM usage (3.65\%-8.46\%) compare to HERA. 
Further removing CD (URC+SSE+SLE) shows a notable increase in accuracy (2.12-3.25\%) but a significant decrease in SLM usage (5.36-8.53\%) compared to w/o SD, highlighting the effectiveness of CD and SD in identifying subtasks suitable for the SLM.
Further removing SLE (URC+SSE) shows a moderate impact, leading to a 3.12-6.54\% 
decrease in SLM usage compared to URC+SSE+SLE and a slight increase in accuracy. 
The most significant impact occurs when SSE is removed and hence other components depending on it are also removed (w/ URC), resulting in accuracy increases of up to 3.56\% and reductions in SLM usage of up to 11.45\% compared to URC+SSE.
This is because leaving only the URC to route tasks results in significantly fewer subtasks assigned to the SLM. 
Based on the results of w/o URC, we see that URC contributes to a moderate increase in SLM usage (3.36-7.98\%) compared to HERA with a minor impact on accuracy, as it provides an initial filter for requests that can be entirely handled by SLM.
These results suggest that each component is crucial for deciding whether a subtask can be moved to the SLM while not compromising accuracy.

\vspace{-0.1in}

\subsection{Performance of Estimators}

Table \ref{tab:estimator-performance} presents the accuracy and latency ratio of HERA's key estimators on both training datasets (HotpotQA, GSM8K) and generalization datasets (DROP, HumanEval, Webshop). 
For the latter, the table presents the accuracy both without and with continual fine-tuning (in parentheses). The continual fine-tuning is executed after every 1000 new requests are processed, allowing the model to adapt to emerging patterns in the data. All estimators demonstrate high accuracy on the training datasets, with accuracy ranging from 81.7\% to 83.8\%. These results indicate that the estimators provide reliable information for HERA's decision-making process. Importantly, when applied to the generalization datasets, the estimators maintain accuracy between 75.8\% and 77.4\%, with small drops. The relatively small drop in accuracy from training to generalization datasets highlights the estimators' ability to adapt to new, unseen queries. These experimental results represent the worst-case scenario, where HERA' key estimators are not continually fine-tuned. With continually fine-tuning using newly received queries, the estimators achieve accuracy between 77.6\% and 81.2\%. The latencies of \emph{subtask predictors} generate relatively higher latency due to their complexity, while the \emph{user request classifier} and \emph{distance predictor} are significantly faster, allowing for efficient decision-making without substantial overhead.  

\begin{table}[h]
\centering
\caption{Performance of HERA's estimators.}
\label{tab:estimator-performance}
\small
\begin{tabular}{lccc}
\hline
\multirow{2}{*}{Component} & \multicolumn{2}{c}{Accuracy (\%)} & Latency \\
\cline{2-3}
 & Training & Generalization & ratio \\
\hline
User Request Classifier & 83.8 & 77.4 (81.2) & 6.7\% \\
Subtask Predictor (SLM) & 83.5 & 75.6 (80.3)  & 45.2\% \\
Subtask Predictor (LLM) & 81.7 & 74.2 (77.6) & 40.8\% \\
Distance Predictor & 83.3 & 75.8 (79.1) & 7.3\% \\
\hline
\end{tabular}
\begin{tablenotes}
\scriptsize
\item Train: HotpotQA, GSM8K; Generalization: DROP, HumanEval, Webshop, MATH
\end{tablenotes}
\end{table}
\begin{figure*}[ht]
    \centering
    \includegraphics[width=0.98\textwidth,height=\textheight,keepaspectratio]{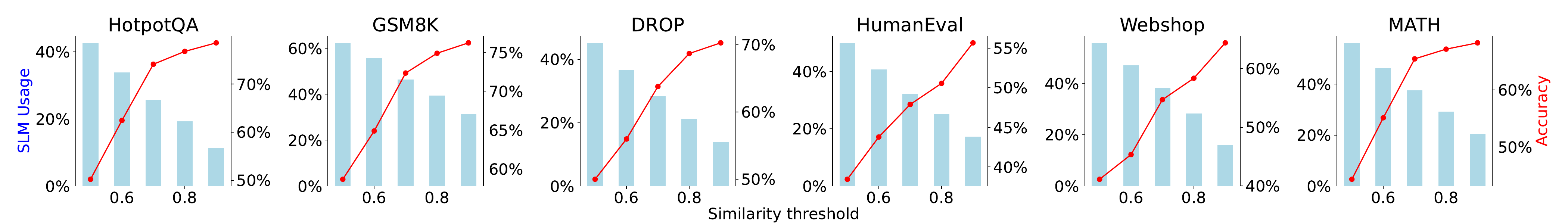}
    \vspace{-0.1in}
    \caption{Sensitivity testing.}
    \vspace{-0.15in}
    \label{fig:sensitivity}
\end{figure*}

\subsection{Sensitivity Testing}
\label{sec: 4.6}


Figure \ref{fig:sensitivity} shows the impact of varying the similarity threshold on HERA's SLM usage and accuracy across six datasets. 
As the similarity threshold increases from 0.5 to 0.9, we observe a consistent trend across all datasets: accuracy improves while SLM usage decreases. For instance, in HotpotQA, the accuracy increases from 50.23\% to 78.56\%, while the SLM usage drops from 42.56\% to 11.23\%. 
GSM8K shows the highest SLM usage (62.23\% to 31.34\%) and HumanEval demonstrates significant accuracy improvement (38.45\% to 55.67\%). These results highlight the trade-off between accuracy and SLM usage in HERA. By analyzing the trade-off ratio between accuracy and SLM usage at different threshold values, one can determine optimal settings for different model pairs and deployment scenarios.

\section{Limitations and Future Work}
\label{sec: 5}
While HERA demonstrates promising results in balancing cost-effectiveness and accuracy for AI agent deployment, several limitations warrant further research:
 
\noindent \textbf{Extensive profiling.} The current system requires extensive profiling of each SLM and LLM pair, which can be time-consuming and hard to retrain. Future work could explore more efficient profiling techniques to leverage information from previously profiled models.



\noindent \textcolor{black}{\textbf{Cost model.} Our current implementation uses a simplified cost model focusing primarily on LLM API costs. However, real-world deployments involve additional considerations: edge device power consumption and hardware costs, data transfer expenses, SLA requirements, hardware utilization, and request queuing patterns. Future work should develop a more comprehensive model incorporating dynamic SLA-aware scheduling, resource utilization monitoring, network-aware decision making, and multi-tenant optimization. These enhancements would better reflect the full range of operational constraints and costs in production environments.}
\noindent \textbf{Multi-model extension.} While HERA currently supports any compatible SLM-LLM pair, as demonstrated with Mistral-7B/GPT-4 and Llama-3.1/Claude 3.5, it could be extended to leverage multiple models simultaneously. This enhancement would enable dynamic model selection based on subtask characteristics and resource constraints, allowing for more efficient task routing across different model capabilities. Implementation would require developing model-specific performance profiles and enhanced resource management strategies to optimize both accuracy and computational efficiency.

\vspace{-0.1in}
\section{Related Work}\label{sec7} 
\textbf{LLM-based AI Agents.} 
Recent developments in LLMs have led to a surge in constructing LLM-based autonomous agents aimed at achieving human-level decision-making capabilities \cite{wei2022chain, brown2020language}. Most studies within this domain can be categorized into three main areas: agent architecture design \cite{qian2023communicative, AutoGPT, wu2023autogen, xu2023rewoo, yao2022react, shinn2023reflexion}, capability acquisition \cite{wang2024survey, schick2024toolformer,shen2024hugginggpt}, and application domains \cite{qian2023communicative, zhu2023ghost, wang2023voyager, bran2023chemcrow}.  These works are complementary to our work and have the potential to be included in HERA.


\noindent \textbf{Hybrid ML Inference.} Recent ML advancements have introduced hybrid inference techniques that strategically combine models of different sizes to optimize cost and efficiency~\cite{kag2023efficient,ding2023hybrid,chen2023frugalgpt,ong2024routellm}. These systems typically route simpler queries to smaller models while directing complex tasks to larger, more capable ones~\cite{kag2023efficient,ding2022efficient, jiang2023llm, chen2023frugalgpt}. While approaches like LLM-Blender~\cite{jiang2023llm} and FrugalGPT~\cite{chen2023frugalgpt} utilize multiple LLMs per request, our method achieves comparable quality with a single LLM invocation, reducing operational overhead. HybridLLM~\cite{ding2023hybrid}, most closely related to our work, routes requests between LLM and SLM based on difficulty, whereas HERA optimizes subtask allocation within an AI agent's decision-making process. Recent research has also explored broader LLM inference optimization challenges like latency, throughput, and resource utilization~\cite{miao2024specinfer,oh2024exegpt,miao2024spotserve,heo2024neupims,sun2024adapipe,yu2022orca,kwon2023efficient}.

\vspace{-0.1in}
\section{Conclusion}\label{sec: 7}
Motivated by the need to optimize the trade-off between LLM inference costs and response accuracy in the AI agent, in this paper, we conducted experiment analysis and made several insightful observations. Based on the observations, we propose HERA, a cost-efficient framework for AI agent in the hybrid cloud-edge environments. HERA addresses the challenge of balancing the accuracy and operational costs of AI agents by leveraging the local-based SLMs and introducing a subtask-level computation partitioning strategy. Extensive experiments on six datasets demonstrate the superior performance of HERA in achieving a good balance between accuracy and cost compared to the state-of-the-art. 



\bibliographystyle{plain}
\bibliography{references}

\begin{thebibliography}{10}

\bibitem{llama3}
Meta AI.
\newblock Llama 3 - github repository.
\newblock \url{https://github.com/meta-llama/llama3}.
\newblock Accessed: 2024-10.

\bibitem{almazrouei2023falcon}
Ebtesam Almazrouei, Hamza Alobeidli, Abdulaziz Alshamsi, Alessandro Cappelli, Ruxandra Cojocaru, M{\'e}rouane Debbah, {\'E}tienne Goffinet, Daniel Hesslow, Julien Launay, Quentin Malartic, et~al.
\newblock The falcon series of open language models.
\newblock {\em arXiv preprint arXiv:2311.16867}, 2023.

\bibitem{anthropic2023claude}
{Anthropic}.
\newblock Claude.
\newblock \url{https://www.anthropic.com}, 2023.
\newblock Version 3.5.

\bibitem{bran2023chemcrow}
Andres~M Bran, Sam Cox, Andrew~D White, and Philippe Schwaller.
\newblock Chemcrow: Augmenting large-language models with chemistry tools.
\newblock {\em arXiv preprint arXiv:2304.05376}, 2023.

\bibitem{brev_llm_cost_estimate}
Brev.
\newblock A bottom-up estimate of llm costs.
\newblock \url{https://brev.dev/blog/llm-cost-estimate}, 2023.
\newblock [Online; accessed Oct-2024].

\bibitem{brown2020language}
Tom Brown, Benjamin Mann, Nick Ryder, Melanie Subbiah, Jared~D Kaplan, Prafulla Dhariwal, Arvind Neelakantan, Pranav Shyam, Girish Sastry, Amanda Askell, et~al.
\newblock Language models are few-shot learners.
\newblock {\em Advances in neural information processing systems}, 33:1877--1901, 2020.

\bibitem{chen2023frugalgpt}
Lingjiao Chen, Matei Zaharia, and James Zou.
\newblock Frugalgpt: How to use large language models while reducing cost and improving performance.
\newblock {\em arXiv preprint arXiv:2305.05176}, 2023.

\bibitem{chen2021evaluating}
Mark Chen, Jerry Tworek, Heewoo Jun, Qiming Yuan, Henrique~Ponde de~Oliveira~Pinto, Jared Kaplan, Harri Edwards, Yuri Burda, Nicholas Joseph, Greg Brockman, Alex Ray, Raul Puri, Gretchen Krueger, Michael Petrov, Heidy Khlaaf, Girish Sastry, Pamela Mishkin, Brooke Chan, Scott Gray, Nick Ryder, Mikhail Pavlov, Alethea Power, Lukasz Kaiser, Mohammad Bavarian, Clemens Winter, Philippe Tillet, Felipe~Petroski Such, Dave Cummings, Matthias Plappert, Fotios Chantzis, Elizabeth Barnes, Ariel Herbert-Voss, William~Hebgen Guss, Alex Nichol, Alex Paino, Nikolas Tezak, Jie Tang, Igor Babuschkin, Suchir Balaji, Shantanu Jain, William Saunders, Christopher Hesse, Andrew~N. Carr, Jan Leike, Josh Achiam, Vedant Misra, Evan Morikawa, Alec Radford, Matthew Knight, Miles Brundage, Mira Murati, Katie Mayer, Peter Welinder, Bob McGrew, Dario Amodei, Sam McCandlish, Ilya Sutskever, and Wojciech Zaremba.
\newblock Evaluating large language models trained on code, 2021.

\bibitem{chen2021codex}
Mark Chen, Jerry Tworek, Heewoo Jun, Qiming Yuan, Henrique~Ponde de~Oliveira~Pinto, Jared Kaplan, Harri Edwards, Yuri Burda, Nicholas Joseph, Greg Brockman, Alex Ray, Raul Puri, Gretchen Krueger, Michael Petrov, Heidy Khlaaf, Girish Sastry, Pamela Mishkin, Brooke Chan, Scott Gray, Nick Ryder, Mikhail Pavlov, Alethea Power, Lukasz Kaiser, Mohammad Bavarian, Clemens Winter, Philippe Tillet, Felipe~Petroski Such, Dave Cummings, Matthias Plappert, Fotios Chantzis, Elizabeth Barnes, Ariel Herbert-Voss, William~Hebgen Guss, Alex Nichol, Alex Paino, Nikolas Tezak, Jie Tang, Igor Babuschkin, Suchir Balaji, Shantanu Jain, William Saunders, Christopher Hesse, Andrew~N. Carr, Jan Leike, Josh Achiam, Vedant Misra, Evan Morikawa, Alec Radford, Matthew Knight, Miles Brundage, Mira Murati, Katie Mayer, Peter Welinder, Bob McGrew, Dario Amodei, Sam McCandlish, Ilya Sutskever, and Wojciech Zaremba.
\newblock Evaluating large language models trained on code.
\newblock 2021.

\bibitem{cobbe2021training}
Karl Cobbe, Vineet Kosaraju, Mohammad Bavarian, Mark Chen, Heewoo Jun, Lukasz Kaiser, Matthias Plappert, Jerry Tworek, Jacob Hilton, Reiichiro Nakano, et~al.
\newblock Training verifiers to solve math word problems.
\newblock {\em arXiv preprint arXiv:2110.14168}, 2021.

\bibitem{ding2022efficient}
Dujian Ding, Sihem Amer-Yahia, and Laks~VS Lakshmanan.
\newblock On efficient approximate queries over machine learning models.
\newblock {\em arXiv preprint arXiv:2206.02845}, 2022.

\bibitem{ding2023hybrid}
Dujian Ding, Ankur Mallick, Chi Wang, Robert Sim, Subhabrata Mukherjee, Victor R{\"u}hle, Laks~VS Lakshmanan, and Ahmed~Hassan Awadallah.
\newblock Hybrid llm: Cost-efficient and quality-aware query routing.
\newblock In {\em The Twelfth International Conference on Learning Representations}, 2023.

\bibitem{dua2019drop}
Dheeru Dua, Yizhong Wang, Pradeep Dasigi, Gabriel Stanovsky, Sameer Singh, and Matt Gardner.
\newblock Drop: A reading comprehension benchmark requiring discrete reasoning over paragraphs.
\newblock {\em arXiv preprint arXiv:1903.00161}, 2019.

\bibitem{fu2023specializing}
Yao Fu, Hao Peng, Litu Ou, Ashish Sabharwal, and Tushar Khot.
\newblock Specializing smaller language models towards multi-step reasoning.
\newblock {\em arXiv preprint arXiv:2301.12726}, 2023.

\bibitem{handler2023balancing}
Thorsten H{\"a}ndler.
\newblock Balancing autonomy and alignment: A multi-dimensional taxonomy for autonomous llm-powered multi-agent architectures.
\newblock {\em arXiv preprint arXiv:2310.03659}, 2023.

\bibitem{he2021debertav3}
Pengcheng He, Jianfeng Gao, and Weizhu Chen.
\newblock Debertav3: Improving deberta using electra-style pre-training with gradient-disentangled embedding sharing, 2021.

\bibitem{hendrycksmath2021}
Dan Hendrycks, Collin Burns, Saurav Kadavath, Akul Arora, Steven Basart, Eric Tang, Dawn Song, and Jacob Steinhardt.
\newblock Measuring mathematical problem solving with the math dataset.
\newblock {\em NeurIPS}, 2021.

\bibitem{heo2024neupims}
Guseul Heo, Sangyeop Lee, Jaehong Cho, Hyunmin Choi, Sanghyeon Lee, Hyungkyu Ham, Gwangsun Kim, Divya Mahajan, and Jongse Park.
\newblock Neupims: Npu-pim heterogeneous acceleration for batched llm inferencing.
\newblock In {\em Proceedings of the 29th ACM International Conference on Architectural Support for Programming Languages and Operating Systems, Volume 3}, pages 722--737, 2024.

\bibitem{hsieh2023distilling}
Cheng-Yu Hsieh, Chun-Liang Li, Chih-Kuan Yeh, Hootan Nakhost, Yasuhisa Fujii, Alexander Ratner, Ranjay Krishna, Chen-Yu Lee, and Tomas Pfister.
\newblock Distilling step-by-step! outperforming larger language models with less training data and smaller model sizes.
\newblock {\em arXiv preprint arXiv:2305.02301}, 2023.

\bibitem{hu2021lora}
Edward~J Hu, Yelong Shen, Phillip Wallis, Zeyuan Allen-Zhu, Yuanzhi Li, Shean Wang, Lu~Wang, and Weizhu Chen.
\newblock Lora: Low-rank adaptation of large language models.
\newblock {\em arXiv preprint arXiv:2106.09685}, 2021.

\bibitem{hu2024automated}
Shengran Hu, Cong Lu, and Jeff Clune.
\newblock Automated design of agentic systems.
\newblock {\em arXiv preprint arXiv:2408.08435}, 2024.

\bibitem{jiang2023mistral}
Albert~Q Jiang, Alexandre Sablayrolles, Arthur Mensch, Chris Bamford, Devendra~Singh Chaplot, Diego de~las Casas, Florian Bressand, Gianna Lengyel, Guillaume Lample, Lucile Saulnier, et~al.
\newblock Mistral 7b.
\newblock {\em arXiv preprint arXiv:2310.06825}, 2023.

\bibitem{jiang2023llm}
Dongfu Jiang, Xiang Ren, and Bill~Yuchen Lin.
\newblock Llm-blender: Ensembling large language models with pairwise ranking and generative fusion.
\newblock {\em arXiv preprint arXiv:2306.02561}, 2023.

\bibitem{kag2023efficient}
Anil Kag and Igor Fedorov.
\newblock Efficient edge inference by selective query.
\newblock In {\em International Conference on Learning Representations}, 2023.

\bibitem{kwon2023efficient}
Woosuk Kwon, Zhuohan Li, Siyuan Zhuang, Ying Sheng, Lianmin Zheng, Cody~Hao Yu, Joseph Gonzalez, Hao Zhang, and Ion Stoica.
\newblock Efficient memory management for large language model serving with pagedattention.
\newblock In {\em Proceedings of the 29th Symposium on Operating Systems Principles}, pages 611--626, 2023.

\bibitem{llama.cpp}
Guillaume Lample.
\newblock llama.cpp: Port of facebook's llama model in c/c++.
\newblock \url{https://github.com/ggerganov/llama.cpp}, 2023.

\bibitem{li2023making}
Yifei Li, Zeqi Lin, Shizhuo Zhang, Qiang Fu, Bei Chen, Jian-Guang Lou, and Weizhu Chen.
\newblock Making language models better reasoners with step-aware verifier.
\newblock In {\em Proceedings of the 61st Annual Meeting of the Association for Computational Linguistics (Volume 1: Long Papers)}, pages 5315--5333, 2023.

\bibitem{liang2023taskmatrix}
Yaobo Liang, Chenfei Wu, Ting Song, Wenshan Wu, Yan Xia, Yu~Liu, Yang Ou, Shuai Lu, Lei Ji, Shaoguang Mao, et~al.
\newblock Taskmatrix. ai: Completing tasks by connecting foundation models with millions of apis.
\newblock {\em arXiv preprint arXiv:2303.16434}, 2023.

\bibitem{lu2023chameleon}
Pan Lu, Baolin Peng, Hao Cheng, Michel Galley, Kai-Wei Chang, Ying~Nian Wu, Song-Chun Zhu, and Jianfeng Gao.
\newblock Chameleon: Plug-and-play compositional reasoning with large language models.
\newblock {\em arXiv preprint arXiv:2304.09842}, 2023.

\bibitem{mialon2023augmented}
Gr{\'e}goire Mialon, Roberto Dess{\`\i}, Maria Lomeli, Christoforos Nalmpantis, Ram Pasunuru, Roberta Raileanu, Baptiste Rozi{\`e}re, Timo Schick, Jane Dwivedi-Yu, Asli Celikyilmaz, et~al.
\newblock Augmented language models: a survey.
\newblock {\em arXiv preprint arXiv:2302.07842}, 2023.

\bibitem{miao2024specinfer}
Xupeng Miao, Gabriele Oliaro, Zhihao Zhang, Xinhao Cheng, Zeyu Wang, Zhengxin Zhang, Rae Ying~Yee Wong, Alan Zhu, Lijie Yang, Xiaoxiang Shi, et~al.
\newblock Specinfer: Accelerating large language model serving with tree-based speculative inference and verification.
\newblock In {\em Proceedings of the 29th ACM International Conference on Architectural Support for Programming Languages and Operating Systems, Volume 3}, pages 932--949, 2024.

\bibitem{miao2024spotserve}
Xupeng Miao, Chunan Shi, Jiangfei Duan, Xiaoli Xi, Dahua Lin, Bin Cui, and Zhihao Jia.
\newblock Spotserve: Serving generative large language models on preemptible instances.
\newblock In {\em Proceedings of the 29th ACM International Conference on Architectural Support for Programming Languages and Operating Systems, Volume 2}, pages 1112--1127, 2024.

\bibitem{nakano2021webgpt}
Reiichiro Nakano, Jacob Hilton, Suchir Balaji, Jeff Wu, Long Ouyang, Christina Kim, Christopher Hesse, Shantanu Jain, Vineet Kosaraju, William Saunders, et~al.
\newblock Webgpt: Browser-assisted question-answering with human feedback.
\newblock {\em arXiv preprint arXiv:2112.09332}, 2021.

\bibitem{Neoteric2023}
Neoteric.
\newblock How much does it cost to use gpt models? gpt-3 pricing explained.
\newblock \url{https://neoteric.eu/blog/how-much-does-it-cost-to-use-gpt-models-gpt-3-pricing-explained/#:~:text=In\%20this\%20scenario\%2C\%20we\%20have,\%2414\%2C4K\%20per\%20month.}, 2023.
\newblock Accessed: Oct, 2024].

\bibitem{oh2024exegpt}
Hyungjun Oh, Kihong Kim, Jaemin Kim, Sungkyun Kim, Junyeol Lee, Du-seong Chang, and Jiwon Seo.
\newblock Exegpt: Constraint-aware resource scheduling for llm inference.
\newblock In {\em Proceedings of the 29th ACM International Conference on Architectural Support for Programming Languages and Operating Systems, Volume 2}, pages 369--384, 2024.

\bibitem{ong2024routellm}
Isaac Ong, Amjad Almahairi, Vincent Wu, Wei-Lin Chiang, Tianhao Wu, Joseph~E Gonzalez, M~Waleed Kadous, and Ion Stoica.
\newblock Routellm: Learning to route llms with preference data.
\newblock {\em arXiv preprint arXiv:2406.18665}, 2024.

\bibitem{ouyang2022training}
Long Ouyang, Jeffrey Wu, Xu~Jiang, Diogo Almeida, Carroll Wainwright, Pamela Mishkin, Chong Zhang, Sandhini Agarwal, Katarina Slama, Alex Ray, et~al.
\newblock Training language models to follow instructions with human feedback.
\newblock {\em Advances in Neural Information Processing Systems}, 35:27730--27744, 2022.

\bibitem{puerto2021metaqa}
Haritz Puerto, G{\"o}zde~G{\"u}l {\c{S}}ahin, and Iryna Gurevych.
\newblock Metaqa: Combining expert agents for multi-skill question answering.
\newblock {\em arXiv preprint arXiv:2112.01922}, 2021.

\bibitem{qian2023communicative}
Chen Qian, Xin Cong, Cheng Yang, Weize Chen, Yusheng Su, Juyuan Xu, Zhiyuan Liu, and Maosong Sun.
\newblock Communicative agents for software development.
\newblock {\em arXiv preprint arXiv:2307.07924}, 2023.

\bibitem{ranaldi2024aligning}
Leonardo Ranaldi and Andre Freitas.
\newblock Aligning large and small language models via chain-of-thought reasoning.
\newblock In {\em Proceedings of the 18th Conference of the European Chapter of the Association for Computational Linguistics (Volume 1: Long Papers)}, pages 1812--1827, 2024.

\bibitem{reimers2019sentence}
Nils Reimers and Iryna Gurevych.
\newblock Sentence-bert: Sentence embeddings using siamese bert-networks.
\newblock {\em arXiv preprint arXiv:1908.10084}, 2019.

\bibitem{schick2024toolformer}
Timo Schick, Jane Dwivedi-Yu, Roberto Dess{\`\i}, Roberta Raileanu, Maria Lomeli, Eric Hambro, Luke Zettlemoyer, Nicola Cancedda, and Thomas Scialom.
\newblock Toolformer: Language models can teach themselves to use tools.
\newblock {\em Advances in Neural Information Processing Systems}, 36, 2024.

\bibitem{shen2024hugginggpt}
Yongliang Shen, Kaitao Song, Xu~Tan, Dongsheng Li, Weiming Lu, and Yueting Zhuang.
\newblock Hugginggpt: Solving ai tasks with chatgpt and its friends in hugging face.
\newblock {\em Advances in Neural Information Processing Systems}, 36, 2024.

\bibitem{shinn2023reflexion}
Noah Shinn, Federico Cassano, Ashwin Gopinath, Karthik~R Narasimhan, and Shunyu Yao.
\newblock Reflexion: Language agents with verbal reinforcement learning.
\newblock In {\em Thirty-seventh Conference on Neural Information Processing Systems}, 2023.

\bibitem{AutoGPT}
Significant-Gravitas.
\newblock Autogpt.
\newblock \url{https://github.com/Significant-Gravitas/AutoGPT}, 2023.

\bibitem{sumers2023cognitive}
Theodore Sumers, Shunyu Yao, Karthik Narasimhan, and Thomas~L Griffiths.
\newblock Cognitive architectures for language agents.
\newblock {\em arXiv preprint arXiv:2309.02427}, 2023.

\bibitem{sun2024adapipe}
Zhenbo Sun, Huanqi Cao, Yuanwei Wang, Guanyu Feng, Shengqi Chen, Haojie Wang, and Wenguang Chen.
\newblock Adapipe: Optimizing pipeline parallelism with adaptive recomputation and partitioning.
\newblock In {\em Proceedings of the 29th ACM International Conference on Architectural Support for Programming Languages and Operating Systems, Volume 3}, pages 86--100, 2024.

\bibitem{exploding_topics_chatgpt_enterprise}
Exploding Topics.
\newblock Chatgpt enterprise: The future of ai in business, 2025.
\newblock Accessed: 2025-01-12.

\bibitem{touvron2023llama}
Hugo Touvron, Thibaut Lavril, Gautier Izacard, Xavier Martinet, Marie-Anne Lachaux, Timoth{\'e}e Lacroix, Baptiste Rozi{\`e}re, Naman Goyal, Eric Hambro, Faisal Azhar, et~al.
\newblock Llama: Open and efficient foundation language models.
\newblock {\em arXiv preprint arXiv:2302.13971}, 2023.

\bibitem{wang2023voyager}
Guanzhi Wang, Yuqi Xie, Yunfan Jiang, Ajay Mandlekar, Chaowei Xiao, Yuke Zhu, Linxi Fan, and Anima Anandkumar.
\newblock Voyager: An open-ended embodied agent with large language models.
\newblock {\em arXiv preprint arXiv:2305.16291}, 2023.

\bibitem{wang2024adapting}
Kuan Wang, Yadong Lu, Michael Santacroce, Yeyun Gong, Chao Zhang, et~al.
\newblock Adapting llm agents with universal feedback in communication.
\newblock In {\em ICML 2024 Workshop on Foundation Models in the Wild}, 2024.

\bibitem{wang2023survey}
Lei Wang, Chen Ma, Xueyang Feng, Zeyu Zhang, Hao Yang, Jingsen Zhang, Zhiyuan Chen, Jiakai Tang, Xu~Chen, Yankai Lin, et~al.
\newblock A survey on large language model based autonomous agents.
\newblock {\em arXiv preprint arXiv:2308.11432}, 2023.

\bibitem{wang2024survey}
Lei Wang, Chen Ma, Xueyang Feng, Zeyu Zhang, Hao Yang, Jingsen Zhang, Zhiyuan Chen, Jiakai Tang, Xu~Chen, Yankai Lin, et~al.
\newblock A survey on large language model based autonomous agents.
\newblock {\em Frontiers of Computer Science}, 18(6):1--26, 2024.

\bibitem{wei2021finetuned}
Jason Wei, Maarten Bosma, Vincent~Y Zhao, Kelvin Guu, Adams~Wei Yu, Brian Lester, Nan Du, Andrew~M Dai, and Quoc~V Le.
\newblock Finetuned language models are zero-shot learners.
\newblock {\em arXiv preprint arXiv:2109.01652}, 2021.

\bibitem{wei2022emergent}
Jason Wei, Yi~Tay, Rishi Bommasani, Colin Raffel, Barret Zoph, Sebastian Borgeaud, Dani Yogatama, Maarten Bosma, Denny Zhou, Donald Metzler, et~al.
\newblock Emergent abilities of large language models.
\newblock {\em arXiv preprint arXiv:2206.07682}, 2022.

\bibitem{wei2022chain}
Jason Wei, Xuezhi Wang, Dale Schuurmans, Maarten Bosma, Fei Xia, Ed~Chi, Quoc~V Le, Denny Zhou, et~al.
\newblock Chain-of-thought prompting elicits reasoning in large language models.
\newblock {\em Advances in Neural Information Processing Systems}, 35:24824--24837, 2022.

\bibitem{wu2023autogen}
Qingyun Wu, Gagan Bansal, Jieyu Zhang, Yiran Wu, Shaokun Zhang, Erkang Zhu, Beibin Li, Li~Jiang, Xiaoyun Zhang, and Chi Wang.
\newblock Autogen: Enabling next-gen llm applications via multi-agent conversation framework.
\newblock {\em arXiv preprint arXiv:2308.08155}, 2023.

\bibitem{xi2023rise}
Zhiheng Xi, Wenxiang Chen, Xin Guo, Wei He, Yiwen Ding, Boyang Hong, Ming Zhang, Junzhe Wang, Senjie Jin, Enyu Zhou, et~al.
\newblock The rise and potential of large language model based agents: A survey.
\newblock {\em arXiv preprint arXiv:2309.07864}, 2023.

\bibitem{xu2023rewoo}
Binfeng Xu, Zhiyuan Peng, Bowen Lei, Subhabrata Mukherjee, Yuchen Liu, and Dongkuan Xu.
\newblock Rewoo: Decoupling reasoning from observations for efficient augmented language models.
\newblock {\em arXiv preprint arXiv:2305.18323}, 2023.

\bibitem{xu2023small}
Canwen Xu, Yichong Xu, Shuohang Wang, Yang Liu, Chenguang Zhu, and Julian McAuley.
\newblock Small models are valuable plug-ins for large language models.
\newblock {\em arXiv preprint arXiv:2305.08848}, 2023.

\bibitem{yang2018hotpotqa}
Zhilin Yang, Peng Qi, Saizheng Zhang, Yoshua Bengio, William~W Cohen, Ruslan Salakhutdinov, and Christopher~D Manning.
\newblock Hotpotqa: A dataset for diverse, explainable multi-hop question answering.
\newblock {\em arXiv preprint arXiv:1809.09600}, 2018.

\bibitem{yao2022webshop}
Shunyu Yao, Howard Chen, John Yang, and Karthik Narasimhan.
\newblock Webshop: Towards scalable real-world web interaction with grounded language agents.
\newblock In {\em ArXiv}, preprint.

\bibitem{yao2022react}
Shunyu Yao, Jeffrey Zhao, Dian Yu, Nan Du, Izhak Shafran, Karthik Narasimhan, and Yuan Cao.
\newblock React: Synergizing reasoning and acting in language models.
\newblock {\em arXiv preprint arXiv:2210.03629}, 2022.

\bibitem{yu2022orca}
Gyeong-In Yu, Joo~Seong Jeong, Geon-Woo Kim, Soojeong Kim, and Byung-Gon Chun.
\newblock Orca: A distributed serving system for $\{$Transformer-Based$\}$ generative models.
\newblock In {\em 16th USENIX Symposium on Operating Systems Design and Implementation (OSDI 22)}, pages 521--538, 2022.

\bibitem{zhang2022opt}
Susan Zhang, Stephen Roller, Naman Goyal, Mikel Artetxe, Moya Chen, Shuohui Chen, Christopher Dewan, Mona Diab, Xian Li, Xi~Victoria Lin, et~al.
\newblock Opt: Open pre-trained transformer language models.
\newblock {\em arXiv preprint arXiv:2205.01068}, 2022.

\bibitem{zhang2019bertscore}
Tianyi Zhang, Varsha Kishore, Felix Wu, Kilian~Q Weinberger, and Yoav Artzi.
\newblock Bertscore: Evaluating text generation with bert.
\newblock {\em arXiv preprint arXiv:1904.09675}, 2019.

\bibitem{zhu2023ghost}
Xizhou Zhu, Yuntao Chen, Hao Tian, Chenxin Tao, Weijie Su, Chenyu Yang, Gao Huang, Bin Li, Lewei Lu, Xiaogang Wang, et~al.
\newblock Ghost in the minecraft: Generally capable agents for open-world enviroments via large language models with text-based knowledge and memory.
\newblock {\em arXiv preprint arXiv:2305.17144}, 2023.

\end{thebibliography}

\end{document}